\documentclass[10pt,journal,compsoc]{IEEEtran}
\ifCLASSOPTIONcompsoc
  \usepackage[nocompress]{cite}
\else
  \usepackage{cite}
\fi
\ifCLASSINFOpdf
\else
\fi
\usepackage{graphicx}
\usepackage{amsmath}
\usepackage{amssymb}
\usepackage{booktabs}
\usepackage{amsthm,amsmath,amssymb}
\usepackage{multirow}
\usepackage{color}
\usepackage{makecell}
\usepackage{mathrsfs}
\usepackage{pifont}
\newcommand{\eg}{\textit{e}.\textit{g}.}
\newcommand{\lyy}[1]{\textcolor[rgb]{1,0,0}{#1}}

\newcommand{\orange}[1]{\textcolor[RGB]{225,166,124}{#1}}
\newcommand{\yellow}[1]{\textcolor[RGB]{235,215,156}{#1}}

\usepackage{xcolor, soul}
\sethlcolor{green}

\begin{document}

%
\title{Explore Contextual Information for 3D Scene Graph Generation}

\author{Yuanyuan Liu, Chengjiang Long, Zhaoxuan Zhang, Bokai Liu, Qiang Zhang, Baocai Yin, and Xin Yang*
\IEEEcompsocitemizethanks{\IEEEcompsocthanksitem 
Yuanyuan Liu, Zhaoxuan Zhang, Bokai Liu, and Xin Yang are with the Department of Electronic Information and Electrical Engineering, Dalian University of Technology, Dalian, 116024, China.
E-mail: Liuyy990415@gmail.com; zhangzx@mail.dlut.edu.cn; liubokai2021@mail.dlut.edu.cn and xinyang@dlut.edu.cn.

\IEEEcompsocthanksitem Chengjiang Long is currently a Research Scientist at Meta Reality Labs, Burlingame, CA, 94010, USA.
E-mail: clong1@meta.com.

\IEEEcompsocthanksitem Qiang Zhang is with the Department of Electronic Information and Electrical Engineering, Dalian University of Technology, Dalian, 116024, China, and also with the Key Lab of Advanced Design and Intelligent Computing (Dalian University), Ministry of Education, 116622, Dalian, China.
E-mail: zhangq@dlut.edu.cn.

\IEEEcompsocthanksitem Baocai Yin is with the Department of Electronic Information and Electrical Engineering, Dalian University of Technology, Dalian, 116024, China, and also with the Beijing Key Laboratory of Multimedia and Intelligent Software Technology, Beijing Institute of Artificial Intelligence, Faculty of Information Technology, Beijing University of Technology, 100124, Beijing, China. E-mail: ybc@dlut.edu.cn.

\IEEEcompsocthanksitem * Xin Yang (xinyang@dlut.edu.cn) is the corresponding authors.

\IEEEcompsocthanksitem Code is available at: https://github.com/YYLiuDLUT/3D\_SCENEGRAPH}

\thanks{Manuscript received April 19, 2005; revised August 26, 2015.}}

\markboth{IEEE Trans Vis Comput Graph,~Vol.~X, No.~X, XXX}%
{Shell \MakeLowercase{\textit{et al.}}: Bare Demo of IEEEtran.cls for Computer Society Journals}
%



\IEEEtitleabstractindextext{%
\begin{abstract}
3D scene graph generation (SGG) has been of high interest in computer vision. Although the accuracy of 3D SGG on coarse classification and single relation label has been gradually improved, the performance of existing works is still far from being perfect for fine-grained and multi-label situations. In this paper, we propose a framework fully exploring contextual information for the 3D SGG task, which attempts to satisfy the requirements of fine-grained entity class, multiple relation labels, and high accuracy simultaneously. Our proposed approach is composed of a Graph Feature Extraction module and a Graph Contextual Reasoning module, achieving appropriate information-redundancy feature extraction, structured organization, and hierarchical inferring. Our approach achieves superior or competitive performance over previous methods on the 3DSSG dataset, especially on the relationship prediction sub-task.

\end{abstract}

}

\maketitle

\IEEEdisplaynontitleabstractindextext

%
\IEEEpeerreviewmaketitle

\IEEEraisesectionheading{\section{Introduction}\label{sec:introduction}}
\IEEEPARstart{T}{he} Scene Graph (SG) organizes the content of a scene into a graph-based representation, which encodes objects as nodes, connected via pairwise relationships as edges, thus representing complex scene knowledge as a compact graphical structure. Scene graph generation (SGG) enables adequate perception and comprehensive understanding of scenes, especially for 3D real-world scenes, and therefore is beneficial to widespread applications, \eg, robot navigation~\cite{lv2020improving}, task planning~\cite{kim20193}, and scene modification and manipulation~\cite{zhou2019scenegraphnet, dhamo2021graph}.

\begin{figure}[t]
	\centering
	\includegraphics [width=0.48\textwidth ]{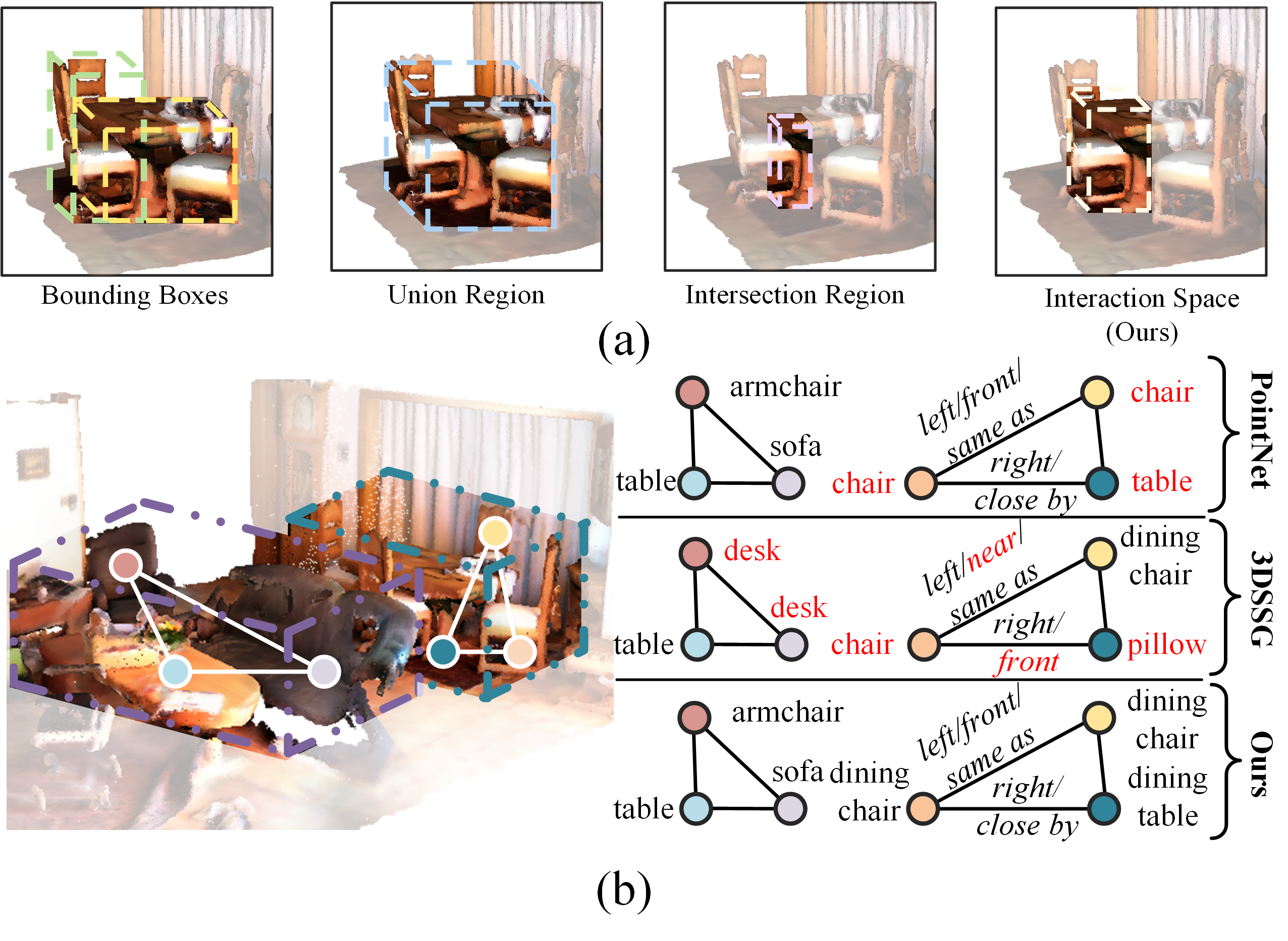}
	\caption{
		\textbf{(a) Different visual patterns for relationship representation.} Corresponding modeling regions for the relationship between a table and a chair. \textbf{(b) Multi-granularity of objects and multi-relationships between objects.} Without the message propagation phase, PointNet cannot infer \texttt{table} as \texttt{dining table}. The full-graph-based message propagation blurs the discrimination of the delicate features, which leads 3DSSG to perform worse than PointNet. With structured organization rules and hierarchical object labels for message propagation, our method accurately predicts fine-grained entities and multiple relation labels. The word marked in \lyy{red} color denotes the wrong prediction.
		}
	\label{fig:teaser}
\end{figure}

One of the main challenges that current models face is the existence of multi-granularity of objects and multi-relationships between objects, as illustrated in Fig.~\ref{fig:teaser}(b). The multi-granularity of objects depends not only on the diverse appearances of objects, but also on the surrounding environment of objects. For example, when chairs surround a \texttt{table}, it most likely becomes a \texttt{dining table}. On the other hand, the interaction between two objects is hard to define with a single relationship, which leads to multi-relationships between objects. For example, there can be both a spatial relationship \texttt{left} and a semantic relationship \texttt{same as} between chairs.

A common approach to address this problem is to define effective visual patterns and collect sufficient contextual information before classifying entities. While a series of SGG methods~\cite{yang2021probabilistic,suhail2021energy,herzig2018mapping,zareian2020bridging,newell2017pixels,zhang2019graphical} have achieved great success in 2D scenes by using the union region as the visual pattern and message passing between all the entities, we cannot apply these 2D methods to 3D scenes directly. The reason is that, compared with 2D image pixels, 3D point clouds are highly unstructured and irregular~\cite{li2021sg}, and due to the inevitable noise in the data, it is more difficult to perform appropriate information extraction and fusion. In a 3D scene, using union regions as visual patterns will lead to noise multiplication. The union region refers to the union of the object bounding boxes, so every time a relationship is modeled, the object region is remodeled (see Fig.~\ref{fig:teaser}(a)). The redundant information generated by the repeated modeling of the identical region confuses the learning ability of the algorithm to a large extent.

In the message passing stage, the features extracted by the visual pattern are used as initial features, which will be propagated indiscriminately among all possible neighbors. The unreasonable information fusion process leads to the explosive propagation of noise. In theory, with a certain depth, all nodes' representations will converge to a stationary point, leading to feature consistency~\cite{cai2021rethinking,wu2020comprehensive}. Resulting in the prediction performance of the fused features being even lower than the original features (see Fig.~\ref{fig:teaser}(b)). These two factors lead to the fact that although the recent 3D SGG works~\cite{wald2020learning, zhang2021exploiting} take extra class-agnostic instance information as input, they still cannot obtain satisfactory results with both fine-grained classification and multiple relation labels.

Based on the observations of humans' scene understanding process. People will adaptively adjust the information redundancy~\cite{dieckmann2007influence} when entering an unknown environment by extracting efficient visual patterns. These visual patterns, as well as environmental contexts, are then structurally organized and hierarchically inferred in the prefrontal cortex (PFC)~\cite{sarafyazd2019hierarchical}. So scene knowledge can be rapidly formed in the brain through optimal processing of small amounts of information.

\begin{figure*}[t]
	\centering
	\includegraphics [width=0.95\textwidth]{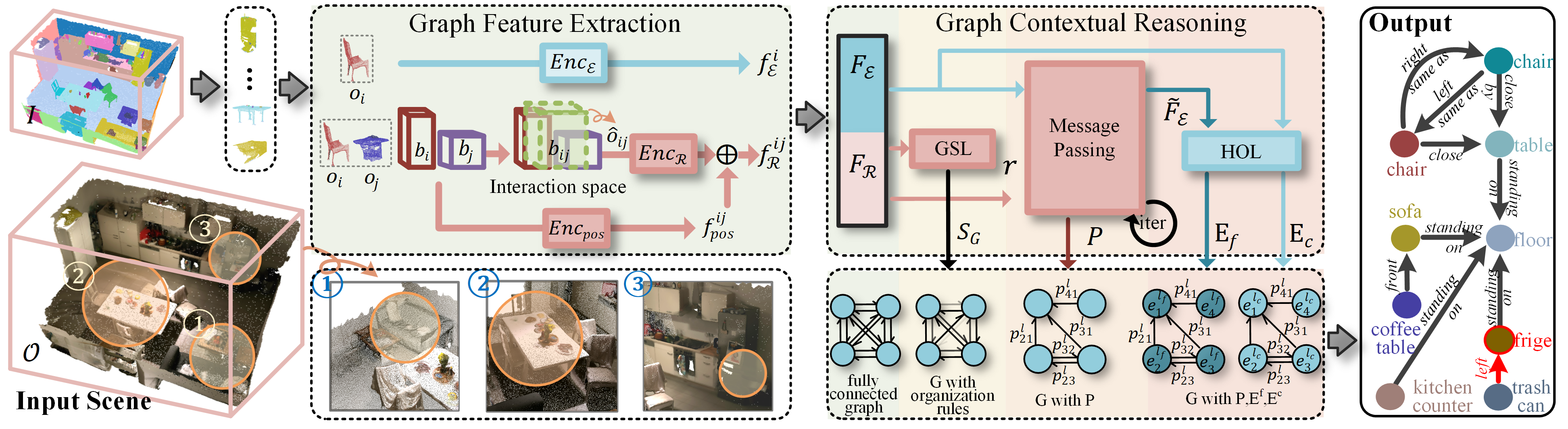}
	\caption{
	\textbf{Overview of our proposed framework.} We only show partial SGG results corresponding to the circularly marked area of the input scene. Our Graph Contextual Reasoning module introduces a Graph Skeleton Learning (GSL) block and a Hierarchy Object Learning (HOL) block on top of the Message Passing block. We differentiate each block with colors and list the evolution of the scene graph as thumbnails below each block. The internal implementation details are shown in Fig.~\ref{fig:contextual}. The generated results demonstrate that our method enables the detection of both \texttt{left} (spatial) and \texttt{same as} (semantic) relationships between two chairs. And it achieves accurate predictions for both coarse and fine object labels, \texttt{table}, and \texttt{coffee table} that belong to the same class but with different granularity. The word marked in \lyy{red} color denotes the wrong prediction.
	}
	\label{fig:pipeline}
\end{figure*}

Inspired by this, we propose a framework for exploring contextual information in 3D SGG, as shown in Fig.~\ref{fig:pipeline}. The Graph Feature Extraction module extracts entity and relation features with appropriate information redundancy. And the Graph Contextual Reasoning module structurally organizes and hierarchically infers these features, as shown in Fig~\ref{fig:contextual}. As a result, we can generate an accurate scene graph with fine-grained entity classes and multiple relation labels.

In the Graph Feature Extraction module, we use the intersection space of the object bounding boxes to replace the union region to reduce the repeated modeling of the identical area. It has been confirmed that the intersection region of the 2D object bounding box proposed by~\cite{wang2019exploring} has a better relationship representation effect than the union region. However, applying it directly to the 3D bounding boxes usually yields inaccurate and unreasonable predictions. To extract rich contextual information between the 3D objects, we present the \textit{``interaction space"} (see Fig.~\ref{fig:interaction}) together with bounding box position information as a new visual pattern for 3D SSG task. In particular, while we reduce the redundancy by adjusting the union region as intersection space, information about objects, relative relationships, and surrounding contextual information is missing in relation features due to the variation of the receptive field. While keeping the information redundancy constant, we recover the over-removed information by region expansion and position information encoding to ensure that the extracted features can cover the underlying properties of the relationship.

In the Graph Contextual Reasoning module, we take a multi-task learning approach by introducing the Graph Skeleton Learning (GSL) block and the Hierarchy Object Learning (HOL) block on top of the Message Passing block. We use graph skeleton information to represent the correlation between object pairs, where graph skeleton information represents the ground truth of the scene graph with the label information removed. The GSL block reconstructs the fully connected graph into an edge-weighted skeleton graph (see Fig.~\ref{fig:skeleton}), and captures the contextual information by propagating node messages in the graph. By jointly training the GSL and message passing blocks, the joint action between the two blocks is strengthened, preventing the GSL blocks from converging independently in static space. GSL utilizes the skeleton to incorporate the structured organization process into the contextual information fusion stage, reducing ineffective information exchange and thus preventing noise propagation.

The GSL block can be regarded as a binary classification task with a simple form and high accuracy. Yet, due to the imbalance of relation and non-relation data among nodes, GSL may block the communication between most nodes. This requires us to use limited contextual information for efficient reasoning to achieve fine-grained classification of entities. Therefore, we design the HOL module, which builds a two-level hierarchical tree using coarse-grained and fine-grained object labels (see in Fig.~\ref{fig:hierarchy}). Coarse-grained object labels supervise initial features that contain only attribute information, and fine-grained object labels supervise features for context fusion. Therefore, HOL reduces the task's difficulty by hierarchically decomposing the object classification task. As demonstrated in our experimental results, the proposed SGG approach significantly outperforms existing state-of-the-art methods quantitatively and qualitatively.

In summary, our contributions are three-fold as follows:
\begin{itemize}
\item We utilize a new visual pattern based on interaction region and bounding box position information to extract relation features with appropriate redundancy.

\item We propose a multi-task learning strategy in our Graph Contextual Reasoning module, which structurally organizes and hierarchically infers the information, predicting multiple labels from similar features. We are the first method to use the graph skeleton information as supervision information.

\item Extensive experiments have demonstrated that our 3D SGG has achieved significant performance improvement to state-of-the-art methods on various scene graph benchmarks, especially on relationship prediction, and even better overcoming the long-tail effect.

\end{itemize}

\section{Related Work}

\noindent\textbf{2D and 3D SGG} Tremendous 2D SGG progress~\cite{lin2020gps,chen2019counterfactual} has been made since~\cite{johnson2015image} firstly mentioned the scene graph. In 3D, SGs have only recently gained more popularity~\cite{armeni20193d, wald2020learning,wu2021scenegraphfusion,zhang2021exploiting,zhang2021knowledge} thanks to the introduction of the 3DSSG dataset~\cite{wald2020learning}, which contains semantically rich scene graphs of 3D scenes.~\cite{wald2020learning} also proposes an end-to-end network that employs a graph convolutional network (GCN)~\cite{kipf2016semi} to handle the message passing stage. SGGPoint~\cite{zhang2021exploiting} builds two associated twin interaction mechanisms between nodes and edges to effectively bridge perception and reasoning. ~\cite{zhang2021knowledge} advocates using graph auto-encoder to automatically extract class-dependent representations and topological patterns as prior knowledge to enhance the accuracy of relationship predictions. Unlike other methods~\cite{gu2019scene,lu2016visual} that use prior knowledge to guide the contextual fusion stage, perception and prior are naively treated as separate components. They are trained separately from different inputs (from images, triples, or label embeddings), and their predictions are usually fused in a probability space. As~\cite{sharifzadeh2021classification}, we train multiple tasks in parallel to achieve better fusion. Our work successfully builds discriminative features based on feature extraction methods and environmental context fusion, enabling us to map between similar features and multiple labels.

\noindent\textbf{Relationship Feature Extraction}~\cite{wald2020learning} uses the same union region as the 2D SGG methods to represent relationship features.~\cite{zhang2021exploiting, zhang2021knowledge} use feature engineering~\cite{wang2019dynamic} and a concatenation scheme to generate relation features from entity features. However, both of them suffer from information redundancy. This redundancy is mainly due to repeated modeling of the same region. Noise in the information is replicated exponentially, reducing features' comprehensibility, ambiguity, and fault tolerance~\cite{dieckmann2007influence}. Most 3D methods try to minimize the noise by reducing the receptive field or locating regions of interest.~\cite{lu2017gpf} attempts point set filtering on point clouds to reconstruct noise-free point sets from corresponding noisy inputs.~\cite{shu2018detecting} learns and predicts interest points in 3D point clouds using multiple feature descriptors. Therefore, we refer to the intersection region~\cite{wang2019exploring} and design a new visual pattern, "interaction space", which has a smaller receptive field and focuses more on the interaction area between objects than the object area. Combined with position feature calculation,  our visual pattern can strike a good balance between information content and redundancy.

\noindent\textbf{Graph Contextual Reasoning} Context modeling strategies~\cite{teng2021target,zareian2020learning,khandelwal2021segmentation,lu2021context,guo2020one} in SGG are mainly used to learn discriminative representation for node and edge prediction, either by designing graph structures or leveraging scene context via various message propagation mechanisms~\cite{zellers2018neural,chiou2021recovering,ren2020scene}. The most popular graph structure is the fully-connected graph~\cite{woo2018linknet,xu2017scene}. Recent works~\cite{yang2019auto,qi2019attentive,yang2018graph} tried to model the context based on the sparse graph structures, using either downstream
tasks (\eg VQA) or trimming functions to cut unessential subject-object pairs. These approaches require additional supervised data~\cite{yao2021visual} to realize the downstream tasks, and the loss of information caused by pruning behavior is not reversible. On the other hand, message propagation mechanisms aim to aggregate the contextual information between entities~\cite{wang2020sketching} and predicates~\cite{cong2020nodis,yin2018zoom,li2018factorizable}. Our work borrows ideas from both sides, which can be seen as incorporating the graph structure into the message propagation process. And like~\cite{zhang2022point}, we design a hierarchy inferring block for object features, thereby reducing the difficulty of object classification and inference tasks through hierarchical decomposition.

\section{Approach}
\subsection{Overview}
Given a scene point cloud $ \mathcal{O}$ labeled with its class-agnostic instance segmentation $\mathcal{I}$ as input, we first split $\mathcal{O}$ into several object point clouds $ \{ o_1, o_2, ..., o_n \}$. Our goal is to predict semantic labels for each object and relations among them. To achieve this, we feed all these objects into Graph Feature Extraction module (Sec.~\ref{SGC}) to obtain entity features $ F_\mathcal{E}$ and relation features $ F_\mathcal{R}$, which form a fully connected feature graph. Then, these two feature sets $ F_\mathcal{E} $ and $ F_\mathcal{R}$ will be structurally organized and hierarchically inferred by Graph Contextual Reasoning module (Sec.~\ref{HCR}), which finally outputs a scene graph denoted as $ \mathcal{G} = (E, P)$. $E$ is a semantic labeled entity set as $ E = \{ e_1^l, e_2^l, ..., e_n^l \}$, and $ P$ is a relation labeled predicate set as $ P = \{ p_{12}^l, p_{13}^l, ..., p_{ij}^l, ..., p_{n,n-1}^l|\ i \neq j\}$. For each relation $ p_{ij}^l$, it is represented as a binary $ m$-dimension vector where $ m$ is the number of all relation labels. If $k_{th} $ dimension's value is 1 means that $ p_{ij}^l$ contains $k_{th} $ relation (All-zero vector represents no relationship between $e_i$ and $e_j$). To be noted that the predicted relations are with multiple relationship labels, so there may be not only one dimension in $ p_{ij}^l $ with a value of 1.

\begin{figure}[t]
	\centering
	\includegraphics[width=0.45\textwidth]{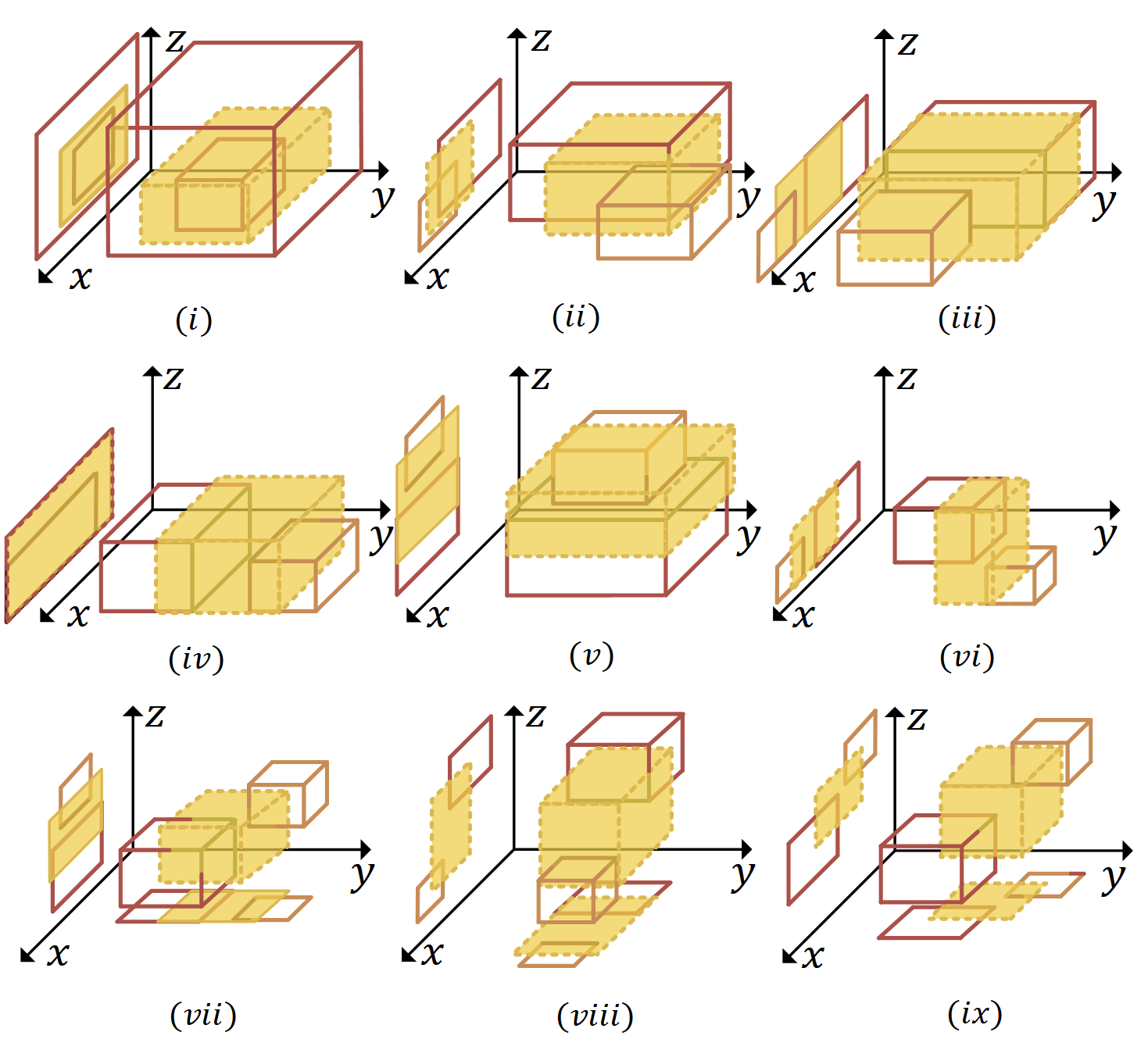} 
	\caption{
	\textbf{Eight cases of interaction space.} Intersectant-($i$) Inclusive, Intersectant-($ii$) Overlap, ($iii$) X-direction disjoint , ($iv$) Y-direction disjoint, ($v$)Z-direction disjoint, ($vi$) XY-direction disjoint, ($vii$) YZ-direction disjoint, ($viii$) XZ-direction disjoint and ($ix$) XYZ-direction disjoint.
	}
	\label{fig:interaction}
\end{figure}

\begin{figure*}[t]
	\centering
	\includegraphics [width=0.95\textwidth]{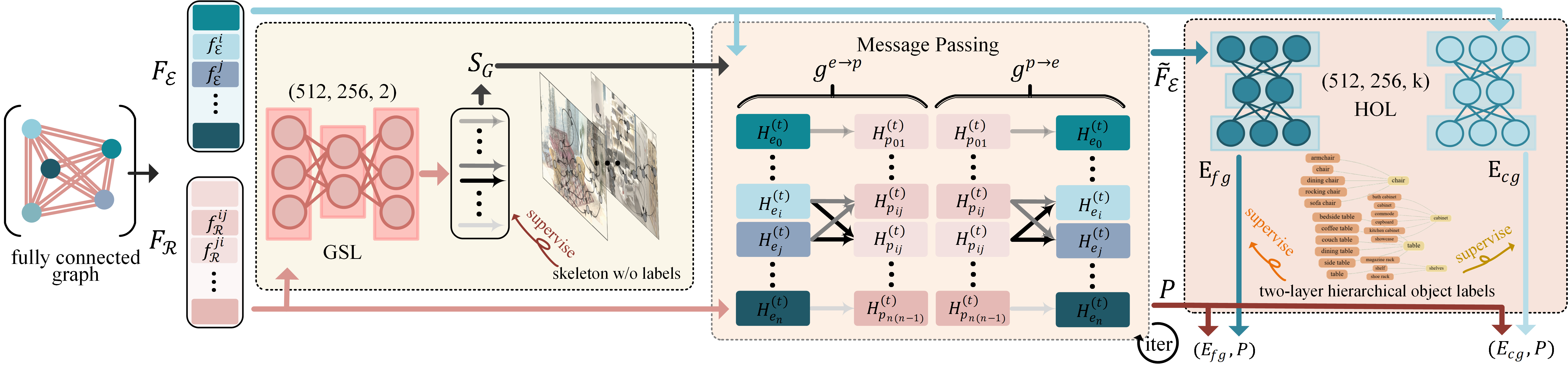}
	\caption{
	\textbf{Graph Contextual Reasoning Module.} The GSL block is supervised by unlabeled graph skeleton information. Coarse and fine-grained entity labels supervise the HOL block. We show part of the labels in the form of a tree structure, where the word with \orange{orange} color indicates the fine-grained class categories, and the word with \yellow{yellow} color indicates the coarse-grained NYU40~\cite{silberman2012indoor} object categories.}
	\label{fig:contextual}
\end{figure*}

\subsection{Graph Feature Extraction}
\label{SGC}
Given a scene point cloud $ \mathcal{O}$, we extract the point set of each instance $i$ separately, labeled with its class-agnostic instance segmentation $\mathcal{I}$, same as~\cite{wald2020learning, wang2019exploring,zhang2021knowledge}. For the object point cloud set $ \{ o_1, o_2, ..., o_n \}$, we first extract the object feature $ f_\mathcal{E}^i$ by a PointNet\cite{qi2017pointnet} based feature extractor $ Enc_\mathcal{E}(\cdot)$ for each object $o_i$, then traverse the following processes for any two objects: 1) calculate the interaction space $B_{ij}$ between bounding box $ b_i$ and $ b_j $ and encode $ b_i$, $ b_j $ to a position feature vector $f_{pos}^{ij}$; 2) extract the features of the points inside $B_{ij}$ by feature extractor $ Enc_\mathcal{R}(\cdot)$ and obtain the relation feature $ f_\mathcal{R}^{ij}$ concatenated with $f_{pos}^{ij}$. The calculation details of interaction space and position feature are described below. 

\noindent\textbf{Interaction Space Calculation} Traditional 3D SGG~\cite{wald2020learning, zhang2021exploiting, zhang2021knowledge} methods represent the relationships that use either union regions or feature engineering. From the perspective of a single pair of objects, these visual patterns contain interaction information, relative position information, surrounding scene information, \textit{etc}, which can cover more contextual information in the relationship features. But looking at the whole scene, the regions of entities are repeatedly modeled by relations, and the relations between several pairs of adjacent objects contain similar or even the same information, resulting in information redundancy. In the subsequent message passing phase, this redundant information will continue to diffuse and grow with iterations, eventually making the features indistinguishable. Therefore, in the feature extraction stage, we design a new visual pattern with reference to the intersection region of~\cite{wang2019exploring}, which is highly coincident with the 3D points in the Interaction Region (IR)~\cite{hu2015interaction} computed using the Interaction Bisector Surfaces (IBS) plane, hence the name Interaction Space (InS). We successfully control the information redundancy by reducing the perceptual regions of relational visual patterns.

To fit the 3D case, we use three auxiliary conditions Eq.~\ref{eq:rp1}, Eq.~\ref{eq:rp2} and Eq.~\ref{eq:rp3} to classify the relative position between $ b_i$ and $b_j$ into eight cases: intersectant (inclusive, overlap), X-direction disjoint, Y-direction disjoint, Z-direction disjoint, XY-direction disjoint, YZ-direction disjoint, XZ-direction disjoint and XYZ-direction disjoint, as shown in Fig.~\ref{fig:interaction}.
\begin{align}
&\left|x_{c}^{i}-x_{c}^{j}\right| \geq \frac{l^{i}+l^{j}}{2}, \label{eq:rp1} \\
&\left|y_{c}^{i}-y_{c}^{j}\right| \geq \frac{w^{i}+w^{j}}{2}, \label{eq:rp2} \\
&\left|z_{c}^{i}-z_{c}^{j}\right| \geq \frac{h^{i}+h^{j}}{2}, \label{eq:rp3}
\end{align}
where $\left(x_{c}^{i}, y_{c}^{i}, z_{c}^{i}\right)$ denote the center coordinates of bounding box $ b_i $, and $l^{i}, w^{i}, h^{i}$ denote $b_i$'s length, width, and height, respectively. The three auxiliary conditions are mainly used to determine whether the projection planes of $ b_i $ and $ b_j $ in the x, y, and z directions intersect.

We first initialize the interaction space $B_{ij}$ with the union bounding box of $b_i$ and $b_j$:

\begin{align}
    B_{ij} = [x_{1}^{U_{ij}},y_{1}^{U_{}ij},z_{1}^{U_{ij}}, x_{2}^{U_{ij}},y_{2}^{U_{ij}},z_{2}^{U_{ij}}]
\end{align}
The lower left corner~$(x_{1}^{U_{ij}},y_{1}^{U_{}ij},z_{1}^{U_{ij}})$ and the upper right corner~$(x_{2}^{U_{ij}},y_{2}^{U_{ij}},z_{2}^{U_{ij}})$ are used to represent the union bounding box of $b_i$ and $b_j$. Judging according to Eq.~\ref{eq:rp1}, Eq.~\ref{eq:rp2} and Eq.~\ref{eq:rp3} in sequence, if a certain auxiliary condition is satisfied, we replace the coordinates of $B_{ij}$ in this direction as below:

\begin{equation}
B_{ij} = \begin{cases}
    x_{1}^{U_{ij}} \rightarrow \min (x_{c}^{i}, x_{c}^{j}), x_{2}^{U_{ij}} \rightarrow \max (x_{c}^{i}, x_{c}^{j}),  & \rm{Eq.\ref{eq:rp1}},\\
    
    y_{1}^{U_{ij}} \rightarrow \min (y_{c}^{i}, y_{c}^{j}), y_{2}^{U_{ij}} \rightarrow \max (y_{c}^{i}, y_{c}^{j}),  & \rm{Eq.\ref{eq:rp2}},\\
    
    z_{1}^{U_{ij}} \rightarrow \min (z_{c}^{i}, z_{c}^{j}), z_{2}^{U_{ij}} \rightarrow \max (x_{c}^{i}, x_{c}^{j}),  & \rm{Eq.\ref{eq:rp3}},
\end{cases}
\end{equation}

For intersectant (inclusive and overlap) case, which no auxiliary condition is met, we first calculate the intersection space of $b_i$ and $b_j$: 
\begin{equation}
\begin{array}{r}
\mathbf{B}_{ij}^{InT}=[\max (x_{1}^{i}, x_{1}^{j}), \max (y_{1}^{i}, y_{1}^{j}), \max (z_{1}^{i}, z_{1}^{j}), \\
\min (x_{2}^{i}, x_{2}^{j}), \min (y_{2}^{i}, y_{2}^{j}), \min (z_{2}^{i}, z_{2}^{j})]
\label{eq:intersectant}
\end{array}
\end{equation}
We use the coordinates of the lower left corner$(x_{1}^{i},y_{1}^{i},z_{1}^{i})$ and the upper right corner$(x_{2}^{i},y_{2}^{i},z_{2}^{i})$ to represent box $b_i$. It is worth mentioning that, as for the intersectant case, we perform region expansion to obtain more environmental context. As the overlap areas of the object bounding boxes are often obscured or inaccessible, there are no or few points in these areas, which makes it difficult to extract relationship features. The interaction space after expansion can be expressed as:
\begin{align}
    B_{ij} = [ \frac{x^{ij}_{1}+x^{U_{ij}}_{1}}{2}, \frac{y^{ij}_{1}+y^{U_{ij}}_{1}}{2}, \frac{z^{ij}_{1}+z^{U_{ij}}_{1}}{2}, \nonumber \\
    \frac{x^{ij}_{2}+x^{U_{ij}}_{2}}{2}, \frac{y^{ij}_{2}+y^{U_{ij}}_{2}}{2}, \frac{z^{ij}_{2}+z^{U_{ij}}_{2}}{2}]
\end{align}
where $\left[x^{ij}_{1},y^{ij}_{1},z^{ij}_{1},x^{ij}_{2},y^{ij}_{2},z^{ij}_{2}\right]$ denotes the intersection bounding box $\mathbf{B}_{ij}^{InT}$ of $b_i$ and $b_j$, calculated by Eq.~\ref{eq:intersectant}.


\noindent\textbf{Position Feature Calculation} The interaction space can effectively reduce information redundancy by narrowing the perception region. However, according to the observation, it can be found that the interaction space is more concerned with the area between objects than the object area. This means that there is a lack of information about objects, relative relationships, and surrounding contextual information in relationship features. Among them, the object and surrounding contextual information can be supplemented in subsequent message passing stages. But the relative relationships are challenging to learn directly from the features due to the disorder of 3D points. Therefore, we need to additionally encode objects' relative positional relationship, which is of great significance for distinguishing the subject-object relationship between objects~\cite{dhingra2021bgt}. Thus we sort the object 3D bounding box coordinates in a subject-object order as~\cite{zhan2019exploring}:
\begin{align}
f_{pos}^{ij} &= Enc_{pos}([p_{min}^{i}, p_{max}^{i}, p_{min}^{j}, p_{max}^{j}]), \\
p_{min}^{i} &= \left[\frac{x_{min}^{b_i}-x_{min}^{U_{ij}}}{l^{U_{ij}}}, \frac{y_{min}^{b_i}-y_{min}^{U_{ij}}}{w^{U_{ij}}}, 
\frac{z_{min}^{b_i}-z_{min}^{U_{ij}}}{h^{U_{ij}}}\right],
\end{align}
where $i$ denotes the subject, $j$ the object, $x_{min}^{b_i}$ the minimum value for $b_i$'s coordinates on X-direction, $x_{min}^{U_{ij}}$ the minimum value for union region's coordinates on X-direction. For brevity, we only report the calculation of $p_{min}^{i}$, the other items are calculated by exchanging $min$ to $max$ or $i$ to $j$. Therefore, we obtain a 12-$d$ bounding box position vector and then extract the position feature $f_{pos}^{ij}$ by a fully connected network $Enc_{pos}(\cdot)$. In general, the feature initialization process for fully connected scene graph can be expressed as:
\begin{align}
F_\mathcal{E}=\{ f_{\mathcal{E}}^{1}, f_{\mathcal{E}}^{2}, ..., f_{\mathcal{E}}^{n}\}&: f_{\mathcal{E}}^{i} = Enc_{\mathcal{E}}\left( o_{i} \right) ,\\
F_\mathcal{R}=\{ f_{\mathcal{R}}^{12}, f_{\mathcal{R}}^{13}, ..., f_{\mathcal{R}}^{n,n-1}\}&:f_{\mathcal{R}}^{ij} = Enc_{\mathcal{R}}(\hat{o}^{ij}) \oplus f_{pos}^{ij},
\end{align}
where $\hat{o}^{ij}$ denotes the points inside intersection space $b_{ij}$, $\oplus$ the concatenation.

\subsection{Contextual Reasoning for Scene Graph Generation}
\label{HCR}
The entity and relation features extracted by the graph feature extraction module form a fully connected graph. Message passing propagates messages through the graph to incorporate contextual information into each node. The contextual information here refers to the underlying dependencies and relations existing in 3D point clouds, which are highly noisy relative to 2D images, so the indiscriminate information communication will aggravate noise propagation. Therefore, we utilize the GSL block to reconstruct the fully connected graph into a skeleton graph, which effectively organizes contextual information by assigning weights to different nodes in the adjacent node set. After the message passing phase, the HOL block establishes long-term connections between initial features and contextual representations by forming coarse-grained and fine-grained labels into a two-level hierarchical tree. We adopt a multi-task learning approach, introduce GSL and HOL blocks on top of the message passing block, and utilize structured organization and hierarchical inferring to achieve context-based reasoning. The implementation details of the three blocks are shown in the Fig.~\ref{fig:contextual}.

In particular, the relation features $F_\mathcal{R}$ are first fed into Graph Skeleton Learning (GSL) block to generate an edge weighted graph skeleton, or in other words a structured organization rule set $\mathcal{S}_G=\{ r_{12}, r_{13}, ..., r_{n,n-1}\}$. Then entity features $F_\mathcal{E}$, relation features $F_\mathcal{R}$ and $\mathcal{S}_G$ are then feed into the Message Passing block, a graph neural network~\cite{scarselli2008graph, li2015gated}, to obtain structurally organized $\tilde{F}_\mathcal{E}$ and $\tilde{F}_\mathcal{R}$. $\tilde{F}_\mathcal{R}$ will then be fed into a predicate predictor to obtain the final predicate set $P$. Finally, we predict the coarse-grained and fine-grained labels for each entity by inputting $F_\mathcal{E}$ and $\tilde{F}_\mathcal{E}$ to Hierarchy Object Learning block. The details of each module are described below.

\begin{figure}[t]
	\centering
	\includegraphics [width=0.48\textwidth]{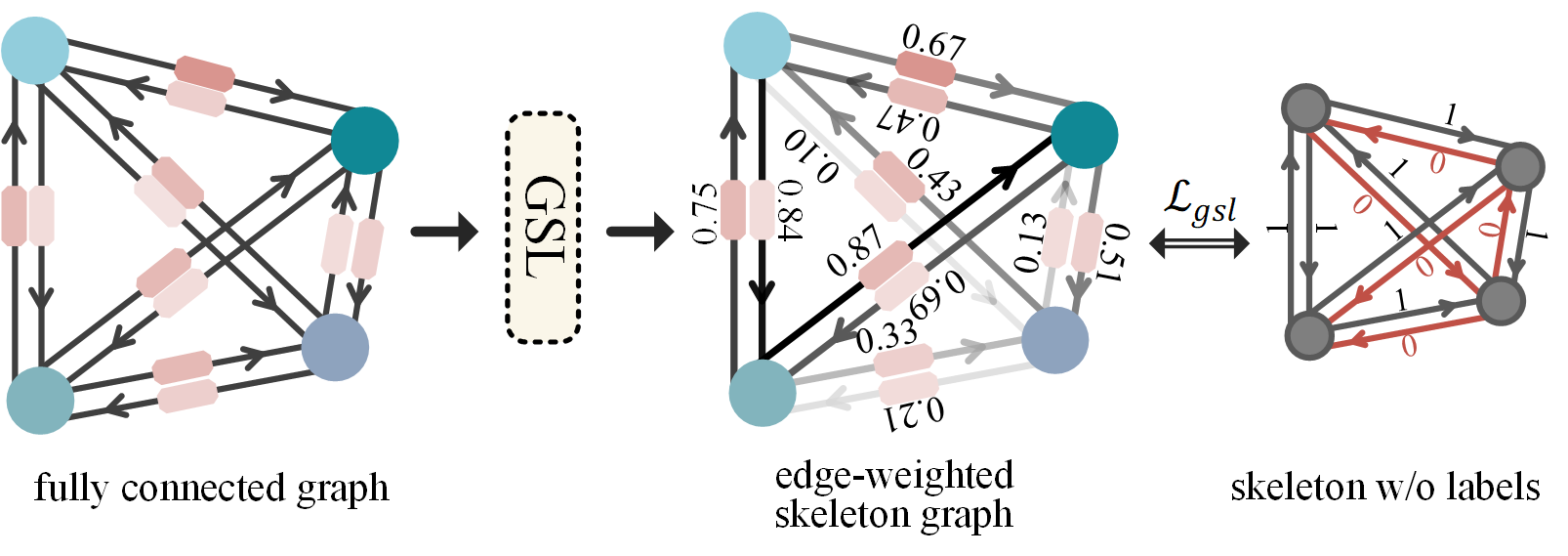}
	\caption{\textbf{Edge-weighted Skeleton Graph.} The graph skeleton information consists of the scene graph ground truth with labels removed. After passing through the GSL block, the edges in the fully connected graph are replaced by the classification confidence. Colors of different edges represent confidence, dark colors represent high confidence, and vice versa. 
	}
\label{fig:skeleton}
\end{figure}

\noindent\textbf{Graph Skeleton Learning} The message passing block implements contextual representation learning by aggregating features between associated nodes. To compute the associations between nodes, we extract the graph skeleton information, which consists of the scene graph ground truth with node and edge labels removed. As shown in Fig.~\ref{fig:skeleton}, under the supervision of the graph skeleton, the GSL block can be viewed as a binary classification task of judging whether there is a relationship between nodes. Its classification confidence can intuitively reflect the associations between nodes. By replacing the edges with the classification confidence, the fully connected graph is reconstructed as an edge-weighted skeleton graph. Our GSL module is implemented by a 3-layer fully connected network with ReLU non-linearity between each layer. It directly takes the initial relation features $F_\mathcal{R}$ as input and defines organizational rules for each edge on the fully connected graph. In particular, for predicate $p_{ij}$ from entity $e_i$ to $e_j$, GSL block takes $f_\mathcal{R}^{ij}$ as input, and predicts its organizational rule $r_{ij}$.  We only take the positive predictive value as the final confidence score and normalize it with $softmax$ option. To achieve hard control for high or low confidence scores, we then feed these scores into the gating function $\tau(\cdot)$ proposed by \cite{li2021bipartite}. This function can make the predicted organization closer to GT scene graph skeleton. The above process can be described as follows:
\begin{align}
 r_{ij}&=\tau\left( softmax(GSL(f_\mathcal{R}^{ij})) \right), \nonumber \\
 \tau(x)&= \left\{\begin{array}{cc}
0 & x \leq \beta \\
\alpha x-\alpha \beta & \beta<x<1 / \alpha+\beta \\
1 & x \geq 1 / \alpha+\beta
\end{array}\right. , \label{eq:gsl} 
\end{align}
where $\alpha$ and $\beta$ are two learnable hyperparameters.

\noindent\textbf{Message Passing}
After computing the organizational rules $\mathcal{S}_G$, we propagate information from each entity to its relevant predicates and vice versa under the guidance of these rules. To this end, we use a multi-stage bipartite graph neural network to propagate the contextual feature on the fully-connected graph within the constraints of organizational rules to get discriminative feature representations. More specifically, the organizational rule $r_{ij}$ affects the information flow of both entity-to-predicate $g^{e \rightarrow p}$ and predicate-to-entity $g^{p \rightarrow e}$. At time step $t$, we denote the hidden state of the entity node $o_i$ as $ H^{(t)}_{e_i} $ and the hidden state of the predicate node $o_{ij}$ as $ H_{p_{i j}}^{(t)} $. We use the feature vector $f_\mathcal{E}^i$ and $f_\mathcal{R}^{ij}$ obtained by the feature extraction module to initialize the hidden state.
\begin{align}
    H^{(0)}_{e_i} = f_\mathcal{E}^i, \quad H^{(0)}_{p_{ij}} = f_\mathcal{R}^{ij}
\end{align}

Our visual pattern narrows the perceptual area to reduce repetitive modeling of the entity area. Therefore, more attention is paid to the area between objects than to the object area. This means that there is a lack of information about objects and surrounding contextual information in relation features. Therefore, this information needs to be supplemented in the message passing phase. Each predicate aggregates messages from its neighbors according to organizational rules, with the formula:
\begin{align}
M_{p_{i j}}^{\left(t\right)}&= g^{e \rightarrow p}(r_{ij}, H^{(t)}_{e_i}, H^{(t)}_{e_j}) \nonumber \\
&=mean\left(g ^ {s} \left(r_{i j} \cdot H_{e_i}^{\left(t\right)}\right)+g^{o}\left(r_{i j} \cdot H_{e_j}^{(t)}\right)\right), \label{equ:e2p}
\end{align}
where $M_{p_{i j}}^{\left(t\right)}$ denotes predicate $p_{ij}$'s message at iteration $t$, $g^{s }$ and $g^{o}$ are independent multi-layer perceptron (MLP) for subject and object, along with a mean operation $mean(\cdot)$.

For each entity node $e_i$, its neighbor nodes will be divided into two categories:  $N_{i*}$ and $N_{*i}$, respectively corresponding to the neighbor nodes where $e_i$ is the subject and $e_i$ is the object. Message from different relationship nodes is aggregated according to their corresponding organizational rules, using the formula:
\begin{align}
M_{e_i}^{\left(t\right)}&=g^{p \rightarrow e}(r_{ij}, r_{ji}, H_{p_{i j}}^{(t)}, H_{p_{ji}}^{(t)}) \nonumber \\
&=mean\left(\sum_{j \in N_{i*}} g^{p}\left(r_{i j} \cdot H_{p_{i j}}^{(t)}\right) \nonumber \right.\\\phantom{=\;\;}
&\left.  +\sum_{j \in N_{*i}(i)} g^{p}\left(r_{j i} \cdot H_{p_{j i}}^{(t)}\right)\right), \label{equ:p2e}
\end{align}
where $M_{e_i}^{\left(t\right)}$ denotes entity $e_i$'s message at iteration $t$, $g^{p}$ is a independent multi-layer perceptron (MLP) for predicate. After the message passing process, we update the feature using two Gated Recurrent Units (GRU)~\cite{cho2014learning}:
\begin{align}
H_{e_{i}}^{(t+1)}&=GRU_{e}\left(H_{e_{i}}^{(t)}, M_{e_{i}}^{(t)}\right), \label{equ:upe}\\ 
H_{p_{i j}}^{(t+1)}&=GRU_{p}\left(H_{p_{i j}}^{(t)}, M_{p_{i j}}^{(t)}\right). \label{equ:upp}
\end{align}
We fuse the information $u$ times to get the final updated contextual-fused object feature $\tilde{F}_\mathcal{E} = \{ H_{e_1}^{(u)}, H_{e_2}^{(u)},...,H_{e_n}^{(u)}\}$ and relation feature $\tilde{F}_\mathcal{R} = \{ H_{p_{12}}^{(u)}, H_{p_{13}}^{(u)},...,H_{p_{n,n-1}}^{(u)}\}$. The organizational rules generated by GSL based on the initial features remain unchanged in the subsequent $u$ fusion processes to prevent the fused relation features from strengthening the association between nodes. After feeding $\tilde{F}_\mathcal{R}$ into the predicate predictor, we obtain the relation labels $ P = \{ p_{12}^l, p_{13}^l, ..., p_{n,n-1}^l \}$.

\begin{figure}[t]
	\centering
	\includegraphics [width=0.48\textwidth]{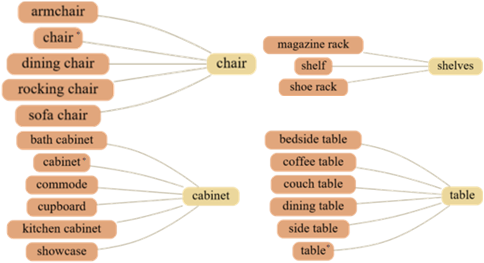}
	\caption{\textbf{Two-layer Hierarchical Tree.} We show part of the labels in the form of a tree structure, where the word marked in \orange{orange} color indicates the fine-grained class categories, and the word marked in \yellow{yellow} color indicates the coarse-grained NYU40~\cite{silberman2012indoor} object categories.}
\label{fig:hierarchy}
\end{figure}

\noindent\textbf{Hierarchy Object Learning} Due to the imbalance of relation and non-relation data, most communication between nodes will be blocked by GSL, and entity nodes can only obtain limited environmental information. Therefore, we utilize the HOL block to perform a hierarchical decomposition of the entity classification task. The HOL block uses coarse-grained and fine-grained entity labels to form a two-level hierarchical tree, as shown in Fig.~\ref{fig:hierarchy}. We construct hierarchical constraints for features containing attribute and contextual information, establishing long-term connections between features and enabling context-based reasoning on objects. The inputs of the HOL block are the initial entity feature $F_\mathcal{E}$ and information-fused feature $\tilde{F}_\mathcal{E}$. There are two entity predictors in HOL block, each entity predictor is implemented by a 3-layer fully connected network with ReLU non-linearity between each layer, which predict coarse and fine entity labels separately:
\begin{align}
    l_{e_i}^c = Dec_\mathcal{E}^{coarse}(f_\mathcal{E}^i), \quad l_{e_i}^f = Dec_\mathcal{E}^{fine}(H_{e_i}^{(u)}).
\end{align}
After labeling $e_i$ with $l_{e_i}^{c}$ and $l_{e_i}^{f}$, we obtain coarse labels $E^c = \{ e_1^{l_c},e_2^{l_c},...e_n^{l_c}\}$ and fine labels $E^f = \{ e_1^{l_f},e_2^{l_f},...e_n^{l_f}\}$. Combined with predicate set $P$, we finally generate the scene graph $\mathcal{G}$ from point cloud $\mathcal{O}$.

\subsection{Loss Functions}

The supervisory signal in the traditional SGG methods is deformed into a triplet form $(e^l_i, p^l_{ij}, e^l_j)$, instead of scene graphs in the intuitive sense, which lacks structure and hierarchy. Under this deformation, the intuitive structured information is transformed into an implicit representation, leading to unsatisfactory performance.
Thus we propose GSL block to generate graph skeleton $ \mathcal{S}_G$ and supervised it with ground truth skeleton $\mathcal{S}_{GT}=\{ b_{12},b_{13},...b_{n,n-1} \}$, where $b_{ij} = 1 $ indicates that there is an edge from $e_i$ to $e_j$ and $b_{ij} = 0 $ the edge does not exist:
\begin{equation}
    \mathcal{L}_{gsl} = BCE(\mathcal{S}_G, \mathcal{S}_{GT}),
\end{equation}
where $BCE(\cdot)$ is a binary cross entropy loss function. Note that our graph skeleton data does not contain any label of nodes and edges, which is intended to limit the model to focus on learning structured information. 

Regarding the HOL block, two granularity object labels were applied as supervision information. Hence, we split the ground truth scene graph into three supervisory signals: 1) skeleton graph; 2) coarse and fine-grained object labels, and 3) predicates. We train our model end-to-end, and our total loss function is described below:
\begin{align}
\mathcal{L}_{total} &= \mathcal{L}_{gsl} + \mathcal{L}_{p} + \lambda _{c}\mathcal{L}_{e}^c + \lambda _{f} \mathcal{L}_{e}^f,
\end{align}
where $\lambda _{c}$ and $\lambda _{f}$ are weighting factors, $\mathcal{L}_{p}$ denotes the per-class binary cross entropy loss, $\mathcal{L}_{e}^c$ and $\mathcal{L}_{e}^f$ are the standard cross entropy loss for multi-class classification task with coarse and fine-grained labels. By jointly training multiple tasks, the modules can be prevented from converging independently in a static space, resulting in better incorporation of structured information and hierarchical labels into visual information.


\newcommand{\tabincell}[2]{\begin{tabular}{@{}#1@{}}#2\end{tabular}}

\subsection{Implementation Details}

We adopt PointNet~\cite{qi2017pointnet} as the feature extraction network, which processes the points inside the object bounding box or interaction space with three channels $\left(x,y,z\right)$ and outputs a final 256-$d$ feature vector. Our message passing block iterates $u = 3$ times. The total number $m$ of relation labels is 26.
In our model, all entity and predicate predictors are composed of three fully connected layers followed by batch normalization and ReLU activation. The $\alpha$, $\beta$ are initialized as 2.2 and 0.025 as suggested by~\cite{li2021bipartite}. The hyperparameters $\lambda_c$, $\lambda_{f}$ in our loss function are both set as 0.1. Adam is chosen as our model optimizer with a learning rate of $10^{-4}$.


\section{Experiments}

\subsection{Dataset and Evaluation Metrics}
3DSSG~\cite{wald2020learning} is a large-scale 3D dataset extended from the 3RSCAN dataset~\cite{wald2019rio} with scene graph annotations. It features 1482 scene graphs, which contain 534 classes of objects and 40 relationships. We take the same 160 object categories and 26 predicate labels as in~\cite{wald2020learning}. 
The dataset provides a variety of object categories with different coarse and fine grains by mapping categories to NYU40~\cite{silberman2012indoor}, RIO27~\cite{wald2019rio} and Eigen~\cite{eigen2014depth}. In this paper, NYU40 is selected as the coarse-grained object label, and the specific mapping relationship is provided in the supplementary material.

For evaluation, we applied the same scene-level split specified in~\cite{wald2020learning} on the point cloud representations. Following~\cite{wu2021scenegraphfusion,zhang2021exploiting, wald2020learning}, the scene graph prediction performance is evaluated upon the three perspectives using the top-$k$ recall metric, namely object class prediction, predicate prediction, and relationship prediction. Of which the relationship level confidence scores are obtained by multiplying each respective score of the subject, predicate, and object in order.

\subsection{Comparison with State-of-the-art Approaches}

We evaluate our proposed method against state-of-the-art scene graph generation methods: MSDN~\cite{li2017scene}, KERN~\cite{chen2019knowledge}, 3DSSG~\cite{wald2020learning}, BGNN~\cite{li2021bipartite}, SGGPoint~\cite{zhang2021exploiting}. In addition to these methods, we also design a simple PointNet~\cite{qi2017pointnet} based method by directly adding the same entity and predicate predictors to justify whether the discrimination of the features will be reduced. Among them, MSDN, KERN, and BGNN are 2D SSG methods. For a fair comparison, we removed the 2D object detector and added the same PointNet based feature extractor as ours (see more comparison results in the supplementary material).

\noindent\textbf{Quantitative Results} As shown in Tab.~\ref{table:result}, our method is the only one that outperforms PointNet on all three sub-tasks without using any prior knowledge. It shows that our defined interaction space minimizes each node's information redundancy, and the contextual reasoning process retains the discrimination of features even after multiple iterations. Our method outperforms PointNet with significant margin of \textbf{6.7} and \textbf{5.1} on object class prediction. Besides, despite the absence of prior statistical co-occurrence knowledge, our method is slightly inferior to KERN by 0.2 and 0.5 on predicate prediction. Ultimately, our method outperforms the others on the relationship prediction sub-task by a large margin.

\begin{table}[t]
\centering
\caption{\textbf{Quantitative comparisons of our method against existing methods on 3DSSG~\cite{wald2020learning} dataset.} \dag~denotes results reproduced with the authors' code.}
\label{table:result}
\setlength\tabcolsep{3pt}
\small
\begin{tabular}{l|ll|ll|ll} 
\hline
\multirow{3}{*}{\textbf{Model}}                                                                                                                                                                                                                                        
                          & \multicolumn{2}{c}{\tabincell{c}{\textbf{Object Class}\\\textbf{Prediction}}} & \multicolumn{2}{c}{\tabincell{c}{\textbf{Predicate}\\\textbf{Prediction}}} & \multicolumn{2}{c}{\tabincell{c}{\textbf{Relationship}\\\textbf{Prediction}}}   \\   
\cmidrule(r){2-3}
\cmidrule(r){4-5}
\cmidrule(r){6-7}

& \textbf{R@5}   & \textbf{R@10}                                                      & \textbf{R@3}   & \textbf{R@5}                                                    & \textbf{R@50}  & \textbf{R@100}                          \\ 
\midrule

PointNet$^{\dag}$~\cite{qi2017pointnet}                 
&63.39  &74.54 &89.07  &96.03  &50.05  &55.73        \\
\midrule
MSDN$^{\dag}$~\cite{li2017scene}
&61.07  &72.41  &85.99  &93.60  &46.55  &53.20  \\
KERN$^{\dag}$~\cite{chen2019knowledge}             
&66.58  &76.52 &\textbf{90.13}  &\textbf{96.61} &51.36  &58.49    \\
3DSSG~\cite{wald2020learning}
&66.41 &77.26  &82.58  &94.34   &51.16  &56.48                        \\
BGNN$^{\dag}$~\cite{li2021bipartite}              
&71.19  &81.98  &86.98  &93.80  &55.20  &60.85   \\
SGGPoint~\cite{zhang2021exploiting}       
&27.82  &35.85  &68.18 &87.32 &7.94  &9.91     \\
\midrule
\textbf{Ours}  
&\textbf{73.40}  &\textbf{82.59}  &89.90  &96.10  &\textbf{61.94}  &\textbf{68.24}   \\
\hline
\end{tabular}
\end{table}

It is worth mentioning that to alleviate the severe object class imbalance problem in the SGG task, SGGPoint only retains 27 object classes and 16 relation classes (3DSSG-O27R16) in the 3DSSG dataset. And it also combines multi-label relationships between nodes into one relationship. Resulting in its network not being able to understand complex scenes with fine-grained objects and multiple relationships well (see our results on 3DSSG-O27R16 in the supplementary material).
\begin{figure}[htbp]
  \centering
  \includegraphics[width=0.95\linewidth]{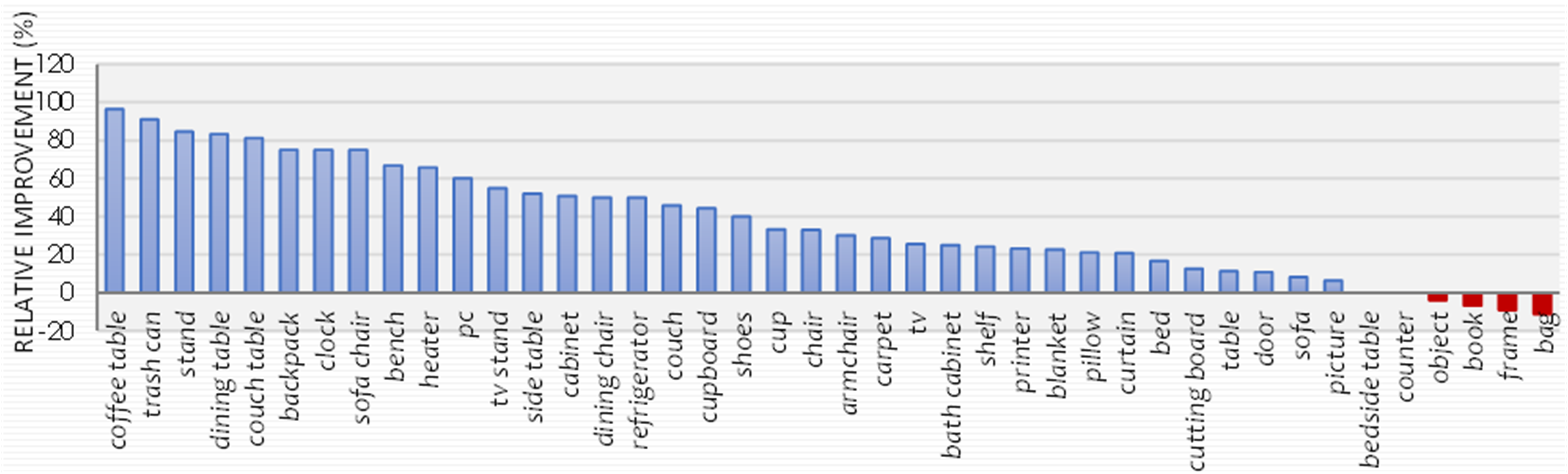}
  \caption{\textbf{Quantitative prediction results about fine-grained object labels.} The R@10 improvement of different entities of our methods to the MSDN.}
  \label{fig:fine_grained}
\end{figure}
Our method can accurately classify entities at a fine-grained level, allowing the construction of scene graphs with more scene knowledge. To more directly compare the performance improvements of fine-grained object classes with MSDN, we further demonstrate the R@10 improvements for some entities in Fig.~\ref{fig:fine_grained}.

It is well-known that SGG models trained on biased datasets have low performances for less frequent categories. Therefore, we additionally introduce mean recall (mR@K) of~\cite{tang2019learning} to examine how well the method learns for uncommon predicates. The metric independently computes the recall for each predicate category and averages the results. So, each category contributes equally. Since our method does not rely on labels or prior knowledge, both of which are biased information, the long-tail effect can be mitigated. As shown in Tab.~\ref{table:mrecall}, our method outperforms all three sub-tasks, effectively reducing the impact of some common but meaningless predicates, such as \texttt{on}, \texttt{near by}, and paying equal attention to those less common predicates, such as \texttt{build in}, \texttt{belonging to}, which are more valuable for high-level reasoning.

\begin{figure}[htbp]
  \centering
  \includegraphics[width=0.95\linewidth]{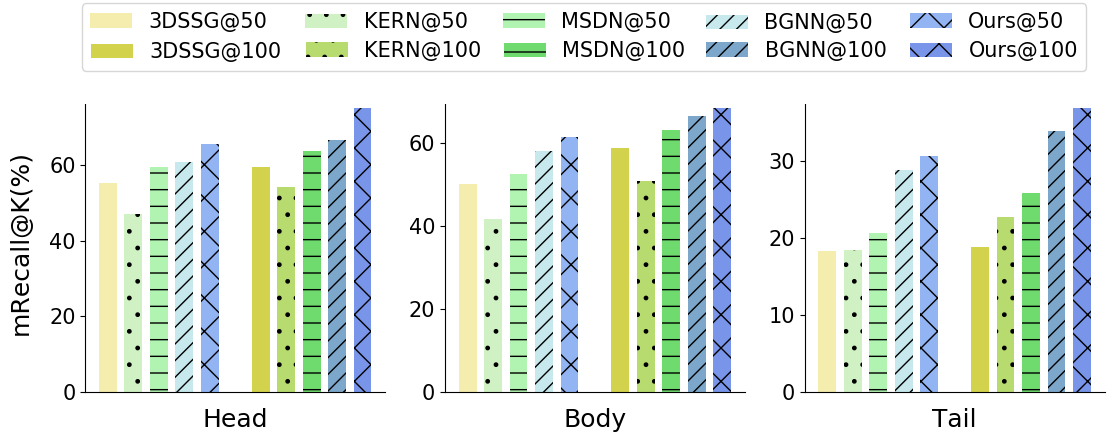}
  \caption{\textbf{The performance of overcoming the long tail effect.} The per-group mR@50 and mR@100 results on each long-tail category groups in relationship prediction subtask.}
  \label{fig:1}
\end{figure}

Moreover, we divided the relationship categories into three disjoint groups according to the
instance number in the training split: head (more than $10^4$), body ($10^3 \sim 10^4$), and tail (less than $10^3$). As shown in Fig.~\ref{fig:1}, we compute the mean recall on each long-tail category group in the relationship prediction sub-task and find our method significantly outperforms the prior works on the tail group. As a result, we achieve the highest mean recall over all categories.

\begin{table}[htbp]
\centering
\caption{\textbf{Quantitative comparisons of our method against existing methods by mR@K.} }
\label{table:mrecall}
\setlength\tabcolsep{2.2pt}
\small
\begin{tabular}{l|ll|ll|ll} 
\hline
\multirow{3}{*}{\textbf{Model}}                                                                                                                                                                                                                                        
                          & \multicolumn{2}{c}{\tabincell{c}{\textbf{Object Class}\\\textbf{Prediction}}} & \multicolumn{2}{c}{\tabincell{c}{\textbf{Predicate}\\\textbf{Prediction}}} & \multicolumn{2}{c}{\tabincell{c}{\textbf{Relationship}\\\textbf{Prediction}}}   \\   
\cmidrule(r){2-3}
\cmidrule(r){4-5}
\cmidrule(r){6-7}

& \textbf{mR@5}   & \textbf{mR@10}                                                      & \textbf{mR@3}   & \textbf{mR@5}                                                    & \textbf{mR@50}  & \textbf{mR@100}                          \\ 
\midrule
MSDN~\cite{li2017scene}                    &23.59  &35.51  &47.41  &62.10  &44.61  &50.17    \\
KERN~\cite{chen2019knowledge}              &23.48  &35.89  &45.68  &61.97  &43.46  &49.14    \\
3DSSG~\cite{wald2020learning}              &23.33  &34.43  &45.82  &63.93  &51.16  &52.21   \\
BGNN~\cite{li2021bipartite}                &28.49  &41.79  &45.15  &58.98  &49.08  &54.21   \\
SGGPoint~\cite{zhang2021exploiting}                &10.54  &12.37  &25.65  &47.59  &1.04  &3.52   \\
\midrule
\textbf{Ours}          &\textbf{33.87}   &\textbf{45.18}     &\textbf{47.10}  &\textbf{64.16}  &\textbf{53.21}   &\textbf{61.50}     \\
\hline
\end{tabular}
\end{table}

\begin{figure*}[htbp]
	\setlength{\tabcolsep}{1pt}\small{
		\begin{tabular}{ccccc}
			
			\includegraphics[width=0.19\linewidth]{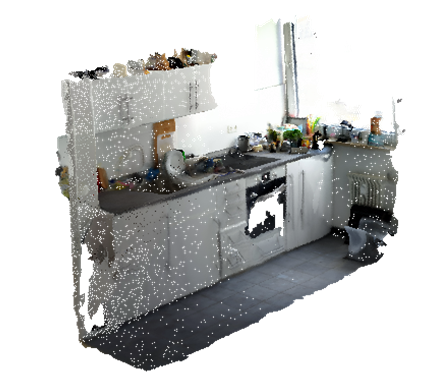}  &
			\includegraphics[width=0.19\linewidth]{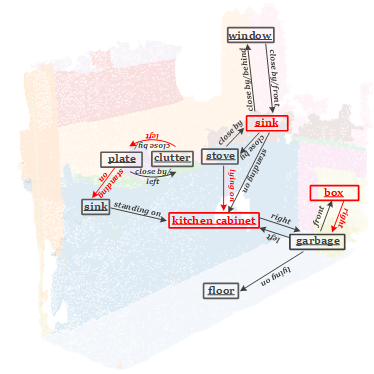}  &
			\includegraphics[width=0.19\linewidth]{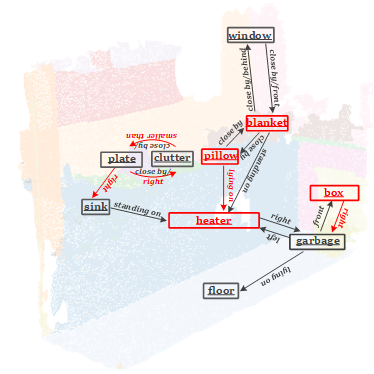} &
			\includegraphics[width=0.19\linewidth]{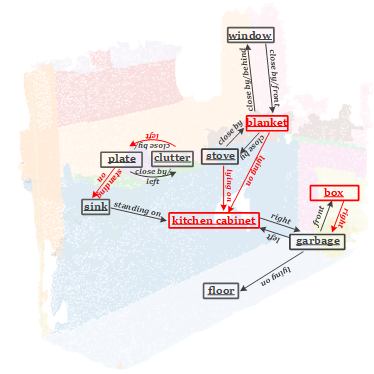} &
			\includegraphics[width=0.19\linewidth]{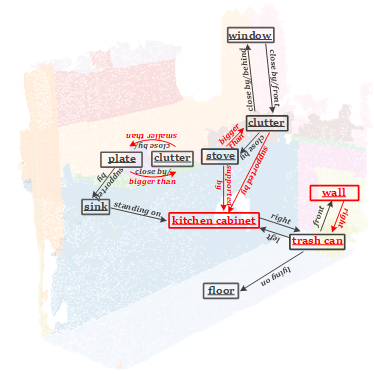}
	        \\
	        Input Point Clouds & PointNet & MSDN & KERN & 3DSSG \\
	        \includegraphics[width=0.19\linewidth]{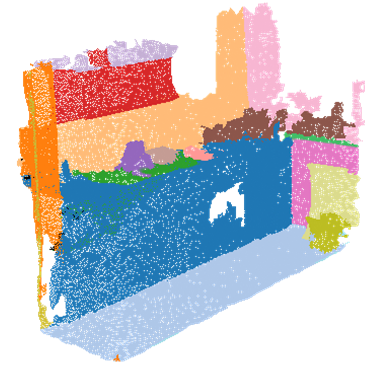}  &
			\includegraphics[width=0.19\linewidth]{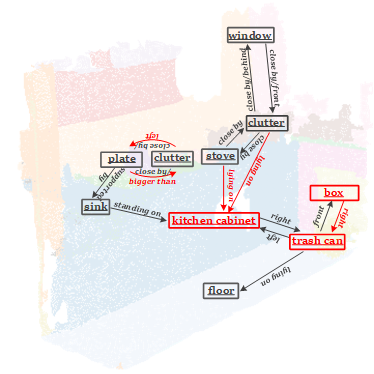}  &
			\includegraphics[width=0.19\linewidth]{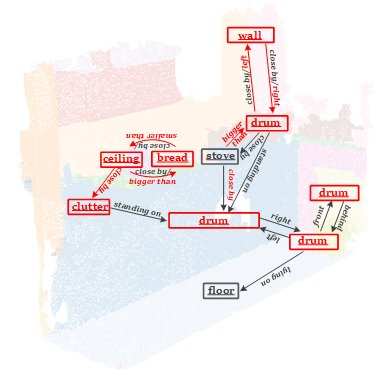} &
			\includegraphics[width=0.19\linewidth]{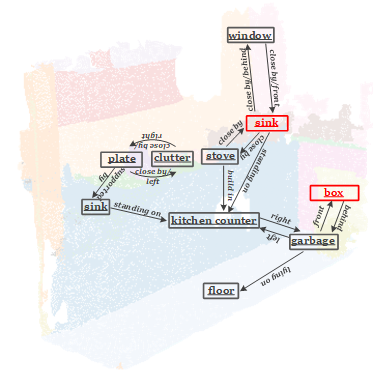} &
			\includegraphics[width=0.19\linewidth]{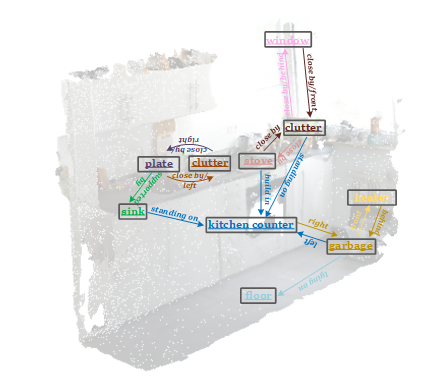} 
	        \\
	        Instance Label & BGNN & SGGPoint & Ours & Ground Truth \\
	        
	        \includegraphics[width=0.19\linewidth]{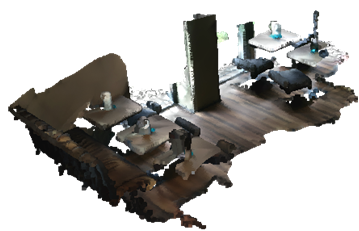}  &
			\includegraphics[width=0.19\linewidth]{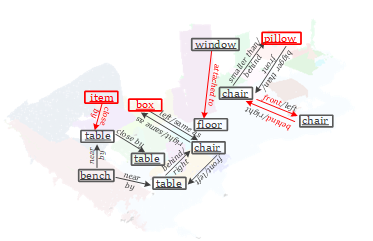}  &
			\includegraphics[width=0.19\linewidth]{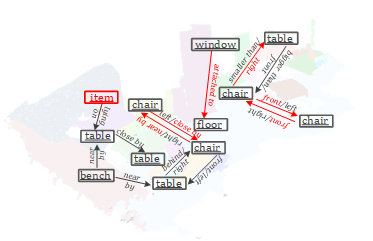} &
			\includegraphics[width=0.19\linewidth]{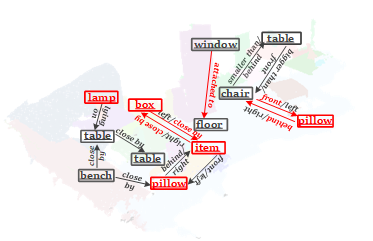} &
			\includegraphics[width=0.19\linewidth]{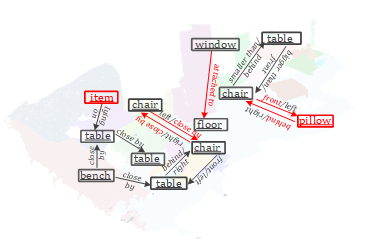}
	        \\
	        Input Point Clouds & PointNet & MSDN & KERN & 3DSSG \\
	        \includegraphics[width=0.19\linewidth]{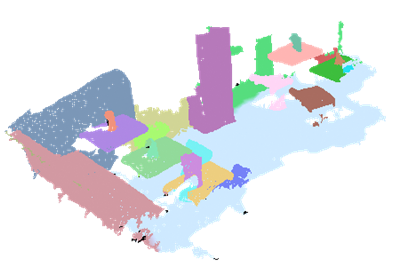}  &
			\includegraphics[width=0.19\linewidth]{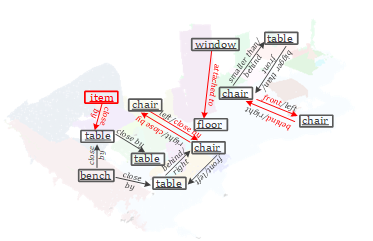}  &
			\includegraphics[width=0.19\linewidth]{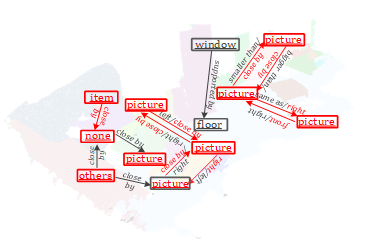} &
			\includegraphics[width=0.19\linewidth]{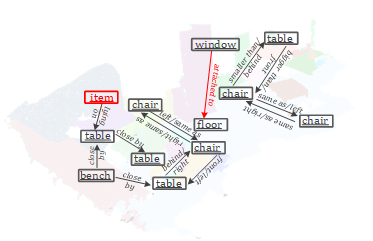} &
			\includegraphics[width=0.19\linewidth]{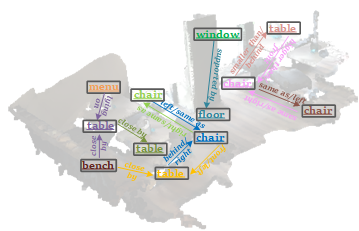} 
	        \\
	        Instance Label & BGNN & SGGPoint & Ours & Ground Truth \\
	        
	        \includegraphics[width=0.19\linewidth]{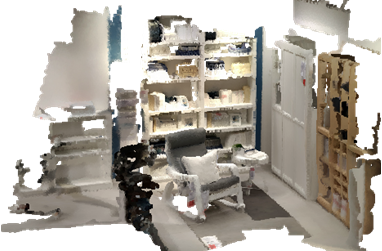}  &
			\includegraphics[width=0.19\linewidth]{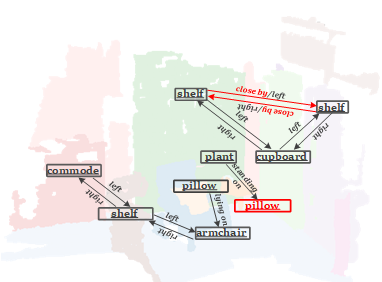}  &
			\includegraphics[width=0.19\linewidth]{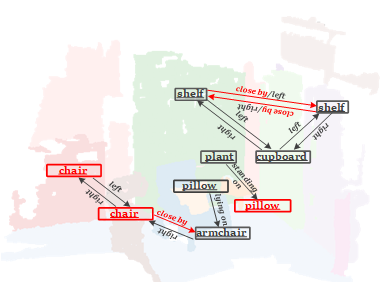} &
			\includegraphics[width=0.19\linewidth]{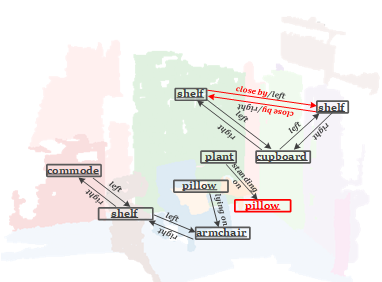} &
			\includegraphics[width=0.19\linewidth]{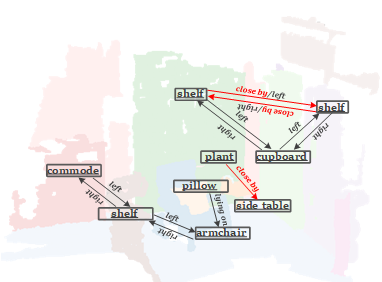}
	        \\
	        Input Point Clouds & PointNet & MSDN & KERN & 3DSSG \\
	        \includegraphics[width=0.19\linewidth]{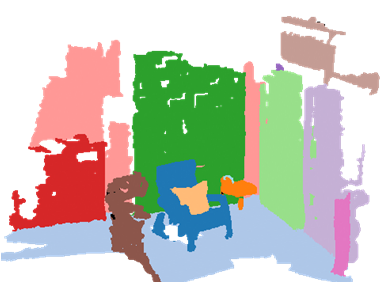}  &
			\includegraphics[width=0.19\linewidth]{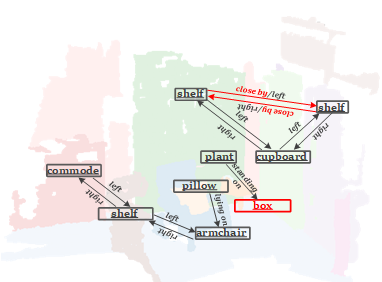}  &
			\includegraphics[width=0.19\linewidth]{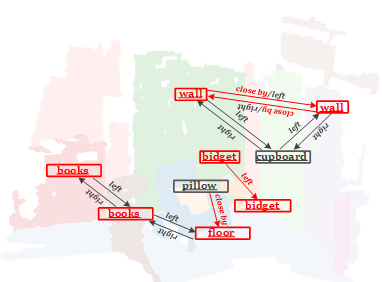} &
			\includegraphics[width=0.19\linewidth]{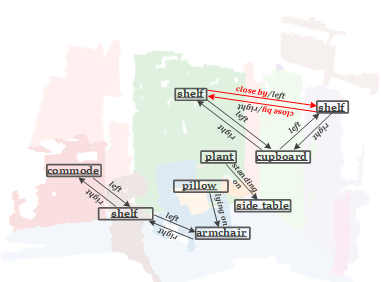} &
			\includegraphics[width=0.19\linewidth]{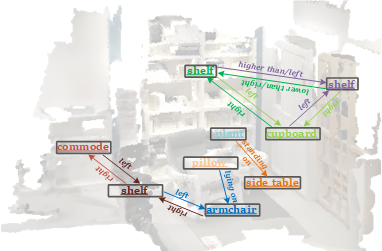} 
	        \\
	        Instance Label & BGNN & SGGPoint & Ours & Ground Truth \\
		
	\end{tabular}}
	
	\caption{\textbf{Comparisons against the state-of-the-arts.} We only show the results for a local area of the scene. We use gray boxes to indicate the entities and underlines to indicate the relations. The different colors of entity names and relation labels are the same as the colors of the class-agnostic instance labels, except that \lyy{red} indicates an incorrect prediction result.}
	
	\label{fig:vis}
\end{figure*}

		

\noindent\textbf{Qualitative Results} As shown in Fig.~\ref{fig:vis}, we chose three categories of indoor scenes, kitchen, cafe, and study room, to verify that the proposed method can accurately explore contextual information to predict scene entities and relationships. In complex environments like kitchens, our method can identify small objects like \texttt{plates} and \texttt{stoves} and accurately distinguish between \texttt{kitchen cabinet} and \texttt{kitchen counter} with similar characteristics. For relations, we successfully predict \texttt{stove-built in-kitchen counter} when other methods only can predict \texttt{lying on}, which requires a deeper semantic understanding of the relationship. 

Our method has a stable prediction for scenes with certain environmental patterns, such as a cafe with multiple sets of tables and chairs. Meanwhile, the prediction result also reflects the drawback of KERN: when the object labels are predicted incorrectly, the relation prediction will be greatly affected, as wrong object labels will bring invalid co-occurrence knowledge to the process of feature information organization. Since KERN relies heavily on prior knowledge, the final SGG results are inconsistent with actual visual information, which is also present in other methods that simply utilize prior knowledge to organize environmental information (see more results in the supplementary materials). Compared with other methods, our result has an excellent performance in predicting the bidirectional relationship between objects. Our method can predict multiple labels accurately from similar features by constructing more discriminative features through effective feature extraction and contextual information fusion (see more qualitative results in the supplementary material). We can also predict incomplete objects in the study scene, such as \texttt{side table} and \texttt{commode}, by supplementing the surrounding contextual information.

\begin{figure*}[htbp]
	\setlength{\tabcolsep}{1pt}\small{
		\begin{tabular}{ccccc}
		
			\includegraphics[width=0.19\linewidth]{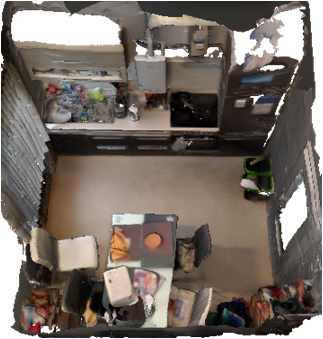} &
			\includegraphics[width=0.19\linewidth]{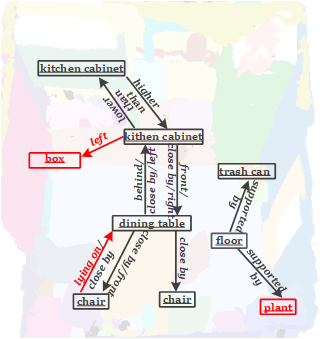} &
			\includegraphics[width=0.19\linewidth]{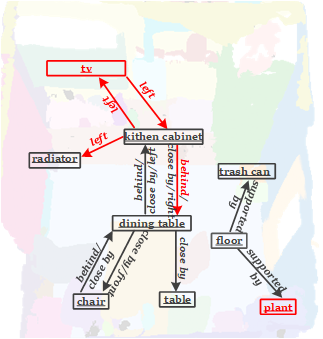} &
			\includegraphics[width=0.19\linewidth]{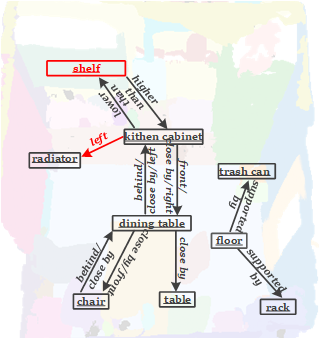} &
			\includegraphics[width=0.19\linewidth]{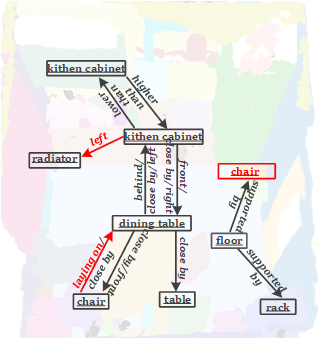}\\
			Input Point Clouds & (a) M0 & (b) M2 & (c) M3 & (d) M4 \\
			
			\includegraphics[width=0.19\linewidth]{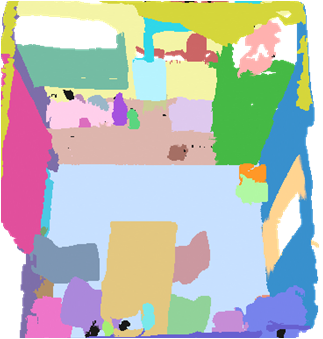} &
			\includegraphics[width=0.19\linewidth]{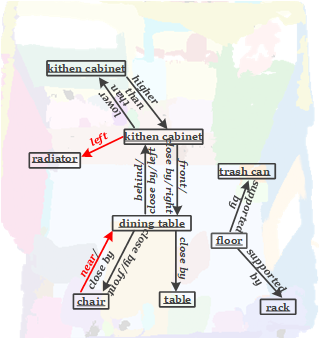} &
			\includegraphics[width=0.19\linewidth]{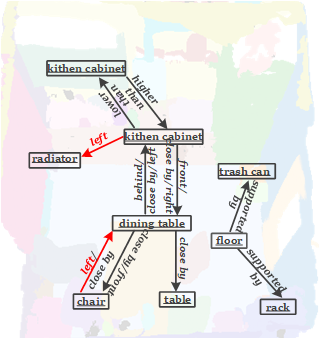} &
			
			\includegraphics[width=0.19\linewidth]{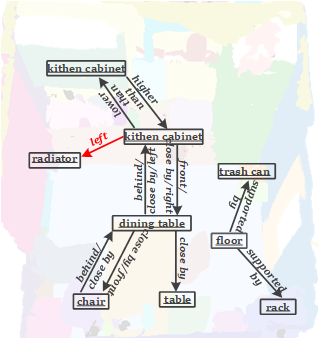} &
			\includegraphics[width=0.19\linewidth]{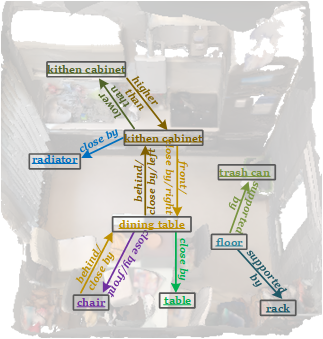}  \\
			Instance Label & (e) M7 & (f) M8 & (g) Ours (M9) & (h) GT \\

	\end{tabular}}
	
	\caption{\textbf{Visualization results for ablation study.} For clarity, we only show the results for a locally representative area of the scene. The rest of the annotation is the same as described above.}
	
	\label{fig:ablation}
\end{figure*}

\begin{table*}[]

\caption{\textbf{Ablation study of our proposed network.} InS: Interaction space; PoF: Position feature; MsP: Message Passing block; GSL: Graph Skeleton Learning block; HOL: Hierarchy Object Learning block. $\rightarrow$\textbf{X} indicates replacing the corresponding component with \textbf{X}. \ding{51} and \ding{55} represent yes and no, respectively. The best results are marked in bold.}
  \centering
\begin{tabular}{l|ll|lll|cc|cc|cc}

\hline
 & \multicolumn{2}{l}{\makecell[c]{Graph Feature \\ Extraction}}\vline &  \multicolumn{3}{l}{\makecell[c]{Graph Contextual \\ Reasoning}}\vline & \multicolumn{2}{l}{\makecell[c]{Object Class \\ Prediction}}\vline                                   & \multicolumn{2}{l}{\makecell[c]{Predicate \\ Prediction}}\vline                                      & \multicolumn{2}{l}{\makecell[c]{Relationship \\ Prediction}}                                   \\
 \cmidrule(r){2-3}
    \cmidrule(r){4-6}
    \cmidrule(r){7-8}
    \cmidrule(r){9-10}
    \cmidrule(r){11-12}
\multirow{-2}{*}{Method} & \makecell[c]{InS}                                & \makecell[c]{PoF}                             & \makecell[c]{MsP}                   & \makecell[c]{GSL}                     & \makecell[c]{HOL}                  & \multicolumn{1}{l}{R@5}  & \multicolumn{1}{l}{R@10}\vline & \multicolumn{1}{l}{R@3}  & \multicolumn{1}{l}{R@5}\vline & \multicolumn{1}{l}{R@50}  & \multicolumn{1}{l}{R@100} \\
\hline 
M0                                               & $\rightarrow$ Union                             & \makecell[c]{\ding{55}}                               & \makecell[c]{\ding{51}}                     & \makecell[c]{\ding{55}}                       & \makecell[c]{\ding{55}}                    & 61.07                                             & 72.41                                             & 85.99                                             & 93.60                                             & 46.55                                             & 53.20                                             \\
M1                                               & $\rightarrow$ InT                               & \makecell[c]{\ding{55}}                               & \makecell[c]{\ding{51}}                     & \makecell[c]{\ding{55}}                       & \makecell[c]{\ding{55}}                    & 61.62 & 73.02 & 83.72 & 92.99 & 46.98 & 53.92 \\
M2                                               & \makecell[c]{\ding{51}}                                  & \makecell[c]{\ding{55}}                              & \makecell[c]{\ding{51}}                     & \makecell[c]{\ding{55}}                       & \makecell[c]{\ding{55}}                    & 63.18                                             & 74.67                                             & 84.96                                             & 93.57                                             & 48.92                                             & 56.25                                             \\
M3                                               & \makecell[c]{\ding{51}}                                  & \makecell[c]{\ding{51}}                               & \makecell[c]{\ding{51}}                     & \makecell[c]{\ding{55}}                       & \makecell[c]{\ding{55}}                    & 69.65                                             & 79.75                                             & 91.41                                             & \textbf{96.70}                                              & 60.92                                             & 66.24                                             \\
\hline
M4                                               & $\rightarrow$ Union                             & \makecell[c]{\ding{55}}                               & \makecell[c]{\ding{51}}                     & \makecell[c]{\ding{51}}                       & \makecell[c]{\ding{55}}                    & 71.97                                             & 81.54                                             & 87.44                                             & 94.26                                             & 58.97                                             & 65.05                                             \\
M5                                               & $\rightarrow$ Union                             & \makecell[c]{\ding{55}}                               & \makecell[c]{\ding{51}}                     & $\rightarrow$ KERN                   & \makecell[c]{\ding{55}}                    & 66.54                                             & 76.53                                             & 89.10                                             & 95.31                                             & 51.32                                             & 57.98                                             \\
M6                                               & $\rightarrow$ Union                             & \makecell[c]{\ding{55}}                               & \makecell[c]{\ding{51}}                     & $\rightarrow$ BGNN                   & \makecell[c]{\ding{55}}                    & 70.10                                             & 80.28                                             & 84.59                                             & 92.40                                             & 54.18                                             & 59.60                                              \\
M7                                               & $\rightarrow$ Union                            & \makecell[c]{\ding{55}}                               & \makecell[c]{\ding{51}}                     & \makecell[c]{\ding{55}}                       & \makecell[c]{\ding{51}}                    & 65.69                                             & 76.44                                             & 86.24                                             & 93.23                                             & 50.32                                             & 56.78                                             \\
M8                                               & $\rightarrow$ Union                             & \makecell[c]{\ding{55}}                               & \makecell[c]{\ding{51}}                     & \makecell[c]{\ding{51}}                       & \makecell[c]{\ding{51}}                    & 72.87                                             & 81.92                                             & 87.62                                             & 94.92                                             & 60.41                                             & 66.16                                             \\
\hline
M9                                               & \makecell[c]{\ding{51}}                                  & \makecell[c]{\ding{51}}                               & \makecell[c]{\ding{51}}                     & \makecell[c]{\ding{51}}                       & \makecell[c]{\ding{51}}                    & \textbf{73.40}                                             & \textbf{82.59}                                            & \textbf{91.43}                                              & 96.49                                              & \textbf{61.94}                                             & \textbf{68.24} \\
\hline
\end{tabular}
\label{table:ablation}
\end{table*}

\subsection{Ablation Study}
\noindent \textbf{Model Components}
As shown in Tab.~\ref{table:ablation}, we first verify the effectiveness of each component by incrementally adding each one of them to a common baseline MSDN~\cite{li2017scene}, denoted as M0. It is worth mentioning that, MSDN uses the union region as the visual pattern, and transfers information indiscriminately between entities and relationships based on a bipartite graph neural network.

 We first make quantitative ablation studies on each component. As we claimed in our paper, repeated modeling and meaningless information communication in the contextual extraction and fusion stages are the main causes for the indistinguishable features. \textbf{InS} reduces the repetitive modeling of the object area by narrowing the perception area (M2: results in object accuracy improved by 2 points). On the other hand, \textbf{PoF} provides InS with information such as the size, location, and subject-predicate relationship of objects that it cannot perceive (M3: results in improved object and predicate accuracy, and relationship prediction task improved for \textbf{13} points). Another source of noise is indiscriminate information fusion. Therefore, \textbf{GSL} reduces meaningless information transmission in the message passing process, and \textbf{HOL} imposes hierarchical constraints on the features before and after fusion (M8: relationship accuracy improved by \textbf{12} points).

\noindent \textbf{Visual Patterns} As shown in Tab.~\ref{table:ablation}, we additionally provide the quantitative comparison results between our interaction space and the intersection region (InT-M1) of~\cite{wang2019exploring}. To fit the data characteristics of 3D point clouds, we expand the interaction space, effectively improving the quality of relation features and the accuracy of relationship prediction by 2 points.

\noindent \textbf{Graph Skeleton Learning Block} Our GSL module provides rules for organizing information at the message passing stage. Moreover, as shown in Tab.~\ref{table:ablation}, our rules can better reduce meaningless information communication compared with the organization rules established by using prior knowledge (KERN-M5) or relationship labels (BGNN-M6), thereby reducing the noise in features.

\noindent \textbf{Qualitative Results} As shown in Fig.~\ref{fig:ablation}, the specific analysis of each case is as follows:

(1) M2: Interaction space is often tiny and covers less environmental context than the union region, which causes lower relationship prediction accuracy. We reduce the information redundancy to improve the accuracy of object prediction. As shown in Fig.~\ref{fig:ablation}$(a)$ and Fig.~\ref{fig:ablation}$(b)$, the entity label \texttt{radiator} has been successfully predicted, but the number of incorrectly predicted relations has also increased.

(2) M3: Because InS cannot perceive object positions and subject-object orders, resulting in the same features on the bidirectional edge between objects. As shown in Fig.~\ref{fig:ablation}$(b)$, M2 predicts the bidirectional edges as \texttt{left} and cannot perceive \texttt{higher than} and \texttt{lower than} relationships because the lack of object positions. By adding PoF, as shown in Fig.~\ref{fig:ablation}$(c)$, the incorrect relation predictions are corrected while maintaining the accuracy of object prediction.

(3) M4: Compared to baseline M0, the additional structured organization rules enable the algorithm to handle the information with high redundancy, even if it has been iterated many times, and still retain sufficient feature discrimination. As shown in Fig.~\ref{fig:ablation}$(a)$ and Fig.~\ref{fig:ablation}$(d)$, structured organization rules improve the prediction results for both entities and relations.

(4) M7: By processing information hierarchically, we realize the process of classifying objects from coarse to fine-grained, and substantially improve object recognition accuracy. As shown in Fig.~\ref{fig:ablation}$(a)$ and Fig.~\ref{fig:ablation}$(e)$, the prediction of entities is significantly improved.

(5) M8: The GSL and HOL blocks together with the Message Passing block form the Graph Contextual Reasoning module, which structurally organizes the information and hierarchically infers the coarse to fine features, as shown in Fig.~\ref{fig:ablation}$(f)$.




\section{Conclusion}
In this work, we propose a framework for the 3D scene graph generation. 
It explores contextual information via a well-designed graph feature extraction module and a graph contextual reasoning module. First, we propose a new visual pattern with appropriate information redundancy. Second, the features extracted from visual patterns are contextually fused by structured organization and hierarchical inferring, retaining the discrimination of features. The experiments demonstrate that the proposed method significantly outperforms the state-of-the-arts methods. In the future, we will try to perform scene understanding for more incomplete scenes through scene completion~\cite{han2019deep} or object reconstruction~\cite{yang2018active}.

\ifCLASSOPTIONcompsoc
  \section*{Acknowledgments}
\else
  \section*{Acknowledgment}
\fi

This work was supported in part by National Key Research and Development Program of China (2022ZD0210500, 2021ZD0112400, 2018AAA0102003), the National Natural Science Foundation of China under Grant  61972067/ U21A20491/U1908214, and the Innovation Technology Funding of Dalian (2020JJ26GX036).


\ifCLASSOPTIONcaptionsoff
  \newpage
\fi



%
{
\bibliographystyle{IEEEtran}
\bibliography{egbib}
}

%

\begin{IEEEbiography}[{\includegraphics[width=1in,height=1.25in,clip,keepaspectratio]{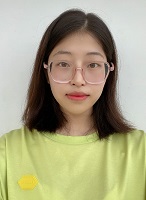}}]{Yuanyuan Liu}
received the B.Eng. degree in the Computer Science and Technology from Northeast Forestry University, Harbin, China. She is currently working towards the Ph.D. degree with the Department of Computer Science and Technology, Dalian, China. Her research interests include scene understanding and computer vision.
\end{IEEEbiography}

\begin{IEEEbiography}[{\includegraphics[width=1in,height=1.25in,clip,keepaspectratio]{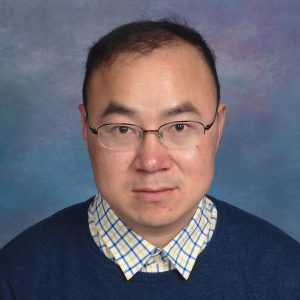}}]{Chengjiang Long} is currently a Research Scientist at Meta Reality Labs. Prior joining Meta, he worked as a Principal Scientist/Tech Leader in JD Tech R\&D Center at Silicon Valley from June 2020 to Dec 2021, and worked as a Computer Vision Researcher/Senior R\&D Engineer at Kitware from February 2016 to April 2020. He also worked as an Adjunct Professor at University at Albany, SUNY from August 2018 to May 2020, and was an Adjunct Professor at Rensselaer Polytechnic Institute (RPI) from Jan 2018 to May 2018. He received the M.S. degree in Computer Science from Wuhan University in 2011 and a B.S degree in Computer Science and Technology from Wuhan University in 2009. He got his Ph.D. degree in Computer Science from Stevens Institute of Technology in 2015. During his Ph.D. study, he worked at NEC Labs America and GE Global Research as a research intern in 2013 and 2015, respectively. To date, he has published over 65 papers including top journals such as TOG, TPAMI, IJCV, TIP and TMM, top international conferences such as SIGGRAPH Asia, CVPR, ICCV, AAAI, and ACM MM, and owns 1 patent. He is also the reviewer for more than 20 top international journals and conferences. His research interests involve various areas of Computer Vision, Computer Graphics, Multimedia, Machine Learning, and Artificial Intelligence. He is a member of IEEE and AAAI.
\end{IEEEbiography}

\begin{IEEEbiography}[{\includegraphics[width=1in,height=1.25in,clip,keepaspectratio]{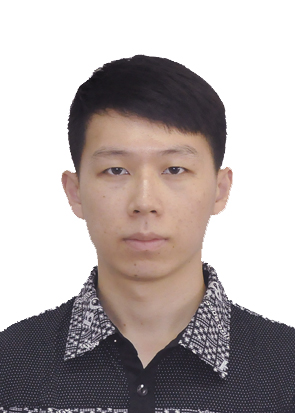}}]{Zhaoxuan Zhang}
	received the B.E. degree in School of Mathematical Sciences from Dalian University of Technology, Liaoning, China in 2016, where he is currently pursuing the Ph.D. degree in computer science. His current research interests include computer vision and computer graphics, especially the 3D reconstruction.
\end{IEEEbiography}

\begin{IEEEbiography}[{\includegraphics[width=1in,height=1.25in,clip,keepaspectratio]{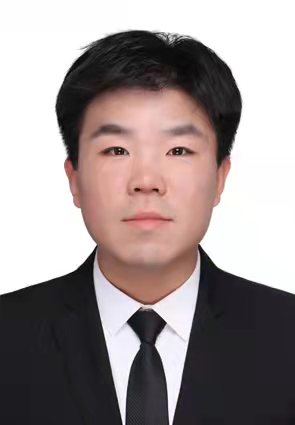}}]{Bokai Liu}
received the B.Eng. degree in the Computer Science and Technology from Taiyuan University of Technology, Taiyuan, China, in 2021. He is currently working toward the M.Sc. degree with the Department of Computer Science and Technology, Dalian University of Technology, Dalian, China. His research interests include scene understanding and computer vision.
\end{IEEEbiography}

\begin{IEEEbiography}[{\includegraphics[width=1in,height=1.25in,clip,keepaspectratio]{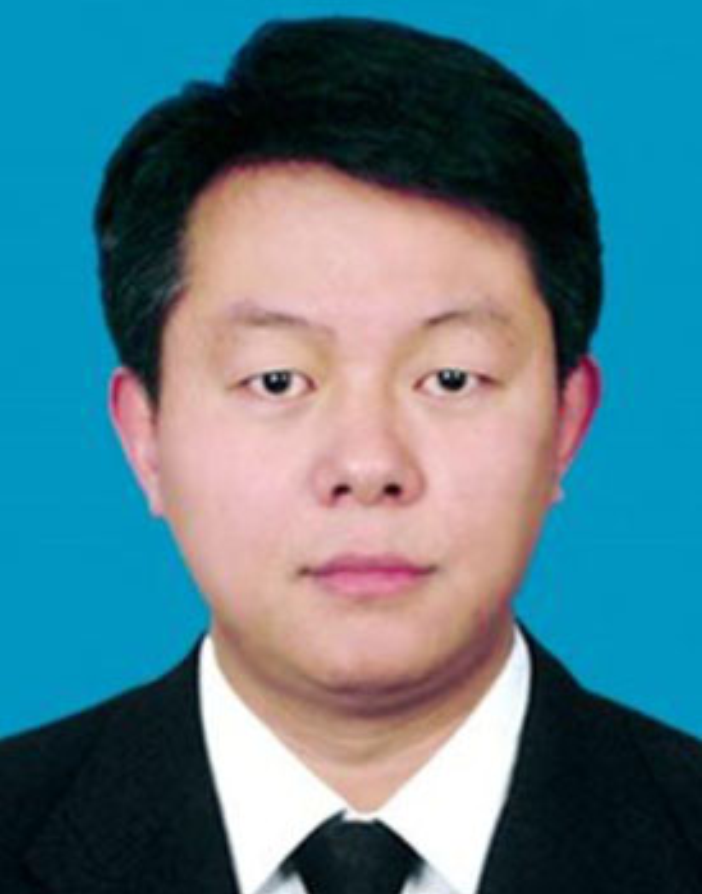}}]{Qiang Zhang}
	was born in Xian, China, in 1971. He received his M.Eng. degree in economic engineering and Ph.D degree in circuits and systems from Xidian University, Xian, China, in 1999 and 2002, respectively. He was a lecturer at the Center of Advanced Design Technology, Dalian University, Dalian, China, in 2003 and was a professor in 2005. His research interests are bio-inspired computing and its applications. He has authored more than 70 papers in the above fields. Thus far, he has served on the editorial board of seven international journals and has edited special issues in journals such as \emph{Neurocomputing} and \emph{International Journal of Computer Applications in Technology}.
\end{IEEEbiography}

\begin{IEEEbiography}[{\includegraphics[width=1in,height=1.25in,clip,keepaspectratio]{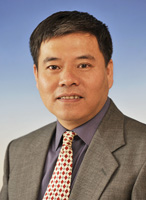}}]{Baocai Yin}
	is a professor and doctoral advisor at Dalian University of Technology. He received his Ph.D. degree in computational mathematics from Dalian University of Technology (1990-1993), where he also received his M.S. degree in computational mathematics (1985-1988) and his B.S. degree in applied mathematics(1981-1985). His research areas include digital multimedia technology, virtual reality and graphics technology, and multi-function perception technology.
\end{IEEEbiography}

\begin{IEEEbiography}[{\includegraphics[width=1in,height=1.25in,clip,keepaspectratio]{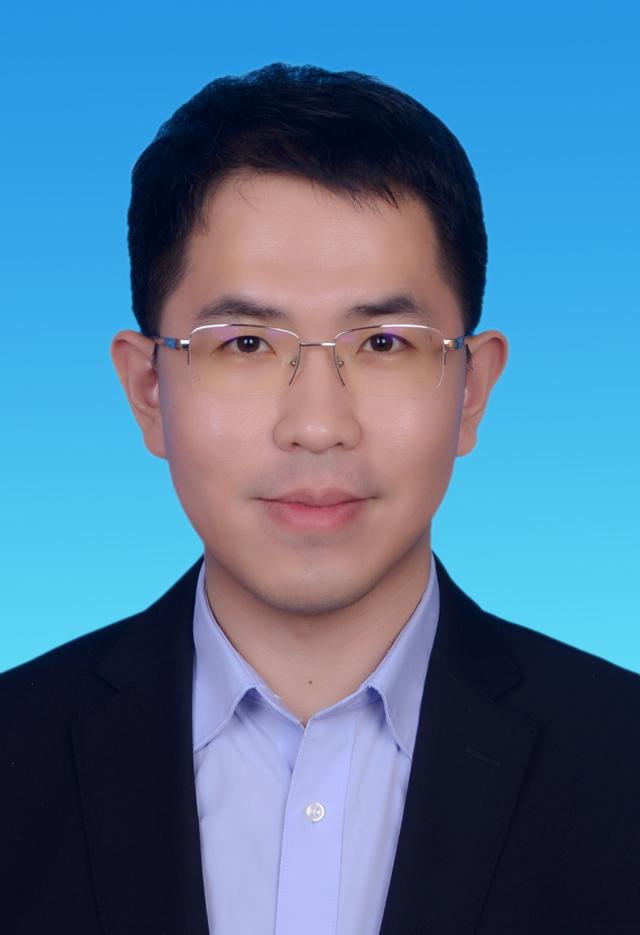}}]{Xin Yang}
	is a professor and doctoral advisor at Dalian University of Technology. He received his Ph.D. degree in computer science from Zhejiang University (2007-2012), and his B.S. degree in computer science from Jilin University (2003-2007). His main research interests include computer graphics and vision, intelligent robot technology, focusing on the efficient expression, understanding, perception and interaction of scenes.
\end{IEEEbiography}




\end{document}